\def\year{2022}\relax
\pdfoutput=1
\documentclass[letterpaper]{article} 
\usepackage{aaai22}  
\usepackage{times}  
\usepackage{helvet}  
\usepackage{courier}  
\usepackage[hyphens]{url}  
\usepackage{graphicx} 
\usepackage{natbib}  
\usepackage{caption} 
\DeclareCaptionStyle{ruled}{labelfont=normalfont,labelsep=colon,strut=off} 
\usepackage{algorithm}
\usepackage{algorithmic}
\usepackage{amsmath}
\usepackage{amsthm}
\usepackage{amssymb}

\usepackage{newfloat}
\usepackage{listings}
\floatstyle{ruled}
\newfloat{listing}{tb}{lst}{}
\floatname{listing}{Listing}
\title{DPVI: A Dynamic-Weight Particle-Based Variational Inference Framework}
\author{
	Chao Zhang$^{1}$, 
	Zhijian Li$^2$\footnote{Chao Zhang and Zhijian Li contribute equally.}, 
	Hui Qian$^{1}$, 
	Xin Du$^2$\footnote{Contact Author.}
}
\affiliations{
	$^1$College of Computer Science and Technology, Zhejiang University\\
	$^2$Information Science and Electronic Engineering, Zhejiang University\\ 
	\{zczju, lizhijian, qianhui, duxin\}@zju.edu.cn
	
}

\usepackage{bibentry}
\usepackage{booktabs}
\usepackage{subfigure}
\usepackage{multirow}
\usepackage{subfiles}
\usepackage[dvipsnames,table,xcdraw]{xcolor}
\usepackage[colorlinks,linkcolor=Blue,anchorcolor=blue,citecolor=Blue,]{hyperref}

\newtheorem{proposition}{Proposition}

\newtheorem{remark}{Remark}

\newcommand{\xb}{\mathbf{x}}

\newcommand{\vb}{\mathbf{v}}
\newcommand{\yb}{\mathbf{y}}

\newcommand{\dr}{\mathrm{d}}
\newcommand{\PM}{\mathcal{P}}

\newcommand{\mut}{\tilde \mu}

\newcommand{\FM}{\mathcal{F}}
\newcommand{\Kb}{\mathbf{K}}
\newcommand{\RBB}{\mathbb{R}}
\begin{document}
	
\maketitle
	
\begin{abstract}
The recently developed Particle-based Variational Inference (ParVI) methods drive the empirical distribution of a set of \emph{fixed-weight} particles towards a given target distribution $\pi$ by iteratively updating particles' positions. However, the fixed weight restriction greatly confines the empirical distribution's approximation ability, especially when the particle number is limited. In this paper, we propose to dynamically adjust particles' weights according to a Fisher-Rao reaction flow. We develop a general Dynamic-weight Particle-based Variational Inference (DPVI) framework according to a novel continuous composite flow, which evolves the positions and weights of particles simultaneously. We show that the mean-field limit of our composite flow is actually a Wasserstein-Fisher-Rao gradient flow of certain dissimilarity functional $\FM$, which leads to a faster decrease of $\FM$ than the Wasserstein gradient flow underlying existing fixed-weight ParVIs. By using different finite-particle approximations in our general framework, we derive several efficient DPVI algorithms. The empirical results demonstrate the superiority of our derived DPVI algorithms over their fixed-weight counterparts.
\end{abstract}
\section{Introduction}
Recently, Particle-based Variational Inference (ParVI) methods have drawn much attention in the Bayesian inference literature, due to their success 
    in efficiently approximating the target posterior distribution $\pi$ 
    \cite{liu2016stein, liu2018riemannian,liu2018stein,pu2017vae,zhu2020variance}.  
The core of ParVIs lies at evolving the empirical distribution of $M$ \emph{fixed-weight} particles by simulating a \emph{continuity equation} 
    through its easy-to-calculate finite-particle position transport approximation 
    \cite{liu2019understanding}. 
Typically, the continuity equation is constructed according to the Wasserstein gradient flow of certain dissimilarity functional $\FM(\mu) := \mathcal D(\mu|\pi)$ 
    vanishing at $\mu = \pi$ \cite{liu2019understanding}.
By using different dissimilarity functionals $\FM$ and position transport approximations, several ParVIs have been proposed, e.g., 
    Stein Variational Gradient Descent (SVGD) method \cite{liu2016stein}, Blob\footnote{This method is originally called $w$-SGLD-B. Here, we follow \cite{liu2019understanding} and denote it as Blob.} \cite{chen2018unified}, 
    GFSD \cite{liu2019understanding}, the Kernel Stein Discrepancy Descent (KSDD) \cite{korba2021kernel}.

\begin{figure}[t]
	\centering
	\includegraphics[width=0.98\linewidth]{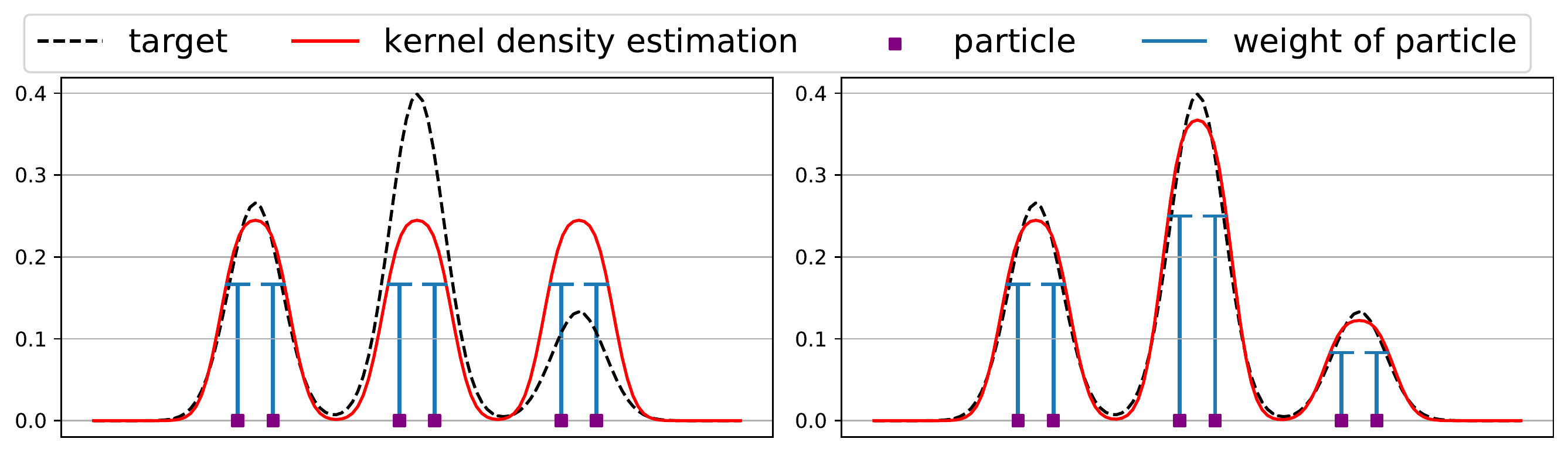}
	\vspace{-1.5mm}
	\caption{\label{introduction}
		Approximating a Gaussian mixture distribution with six particles. 
		The black dashed lines denote the target density, the solid red lines are the densities of particles (estimated using the kernel density estimator), 
		and the heights of the solid blue lines represent the weight of each particle. 
		The left sub-figure shows the result of the fixed-weight ParVI algorithm Blob, 
		and the right sub-figure is from our dynamic-weight D-Blob-CA algorithm.}
	\vspace{-4.5mm}
\end{figure}

\noindent{\bf Fixed weight restriction.}
Existing ParVIs have a common fixed weight restriction, i.e., they all keep the particles' weights fixed during the whole procedure, 
    and only update the positions of particles according to the position transport approximation derived from a continuity equation.
This restriction severely confines the empirical distribution's approximation ability, especially when the particle number $M$ is limited 
    (depicted in Figure \ref{introduction}). 
To mitigate the influence of this restriction, 
    existing ParVIs require plenty of particles to obtain satisfying approximation accuracy \cite{liu2018stein,korba2020non}.
As a result, a large amount of computation is usually needed since the per-iteration computational cost in ParVIs is 
    typically in the square order of $M$.
Actually, a huge deviation between the empirical distribution and the target $\pi$ is often observed when the particle number is insufficient due to a 
    limited computational budget \cite{zhang2020stochastic, zhang2020variance}.

Thus, it is in great need to find an effective weight adjustment strategy 
and design dynamic-weight ParVI algorithms which could achieve a high approximation accuracy with fewer particles and hence less computation. 
Note that, though the continuity equation underlying existing fixed-weight ParVIs can be directly transformed into an easy-to-calculate position transport approximation, 
    adjusting weights according to the continuity equation is generally infeasible as the divergence operator in it would introduce great computational challenge.
Constructing effective algorithms to evolve a set of 
    dynamic-weight particles towards the target $\pi$ is still an open problem in the ParVI field.

To tackle this problem, we propose to dynamically adjust particles' weights according to the Fisher-Rao reaction flow of the underlying functional $\FM$,
    and design a continuous-time composite flow, which evolves the positions and weights of $M$ particles simultaneously.
Specifically, the composite flow is a combination of a finite-particle approximation of the reaction flow and
    a finite-particle position transport approximation of the continuity equation.
Different dynamic-weight ParVI algorithms can be obtained 
by discretizing the continuous flow with different discretization schemes and dissimilarity functionals $\FM$.
The contribution of our paper are listed as follows:
\begin{itemize}
    \item 
    We show that the mean-field limit of our proposed composite flow is actually the gradient flow
    of $\FM$ in the Wasserstein-Fisher-Rao space, which leads to a faster decrease of $\FM$ compared with both the Wasserstein gradient flow
    underlying existing fixed-weight ParVI algorithms and the Fisher-Rao reaction flow.
    \item 
    We propose a general Dynamic-weight Particle-based Variational Inference (DPVI) framework, 
        which utilizes an Euler discretization of the composite flow and adopts a Gauss-Siedel-type strategy to update the positions and the weights. 
    Note that the weight adjustment step can be implemented without bringing much extra computation compared to the position update step.
    By adopting different dissimilarity functionals $\mathcal{F}$ in DPVI,
        we can obtain different efficient dynamic-weight ParVI algorithms.
    \item We propose three efficient DPVI algorithms 
        by using different dissimilarity functionals $\FM$ and their associated finite-particle approximations in the general framework.
    Besides, we also derive three duplicate/kill variants of our proposed algorithms, 
    where a probabilistic discretization to the weight adjustment part in the composite flow is used to 
        dynamically duplicate/kill particles, instead of adjusting the particles' weights continuously.
\end{itemize}
We evaluate our algorithms on various synthetic and real-world tasks.
The empirical results demonstrate the superiority of our dynamic weight strategy over the fixed weight strategy,  
and our DPVI algorithms constantly outperform their fixed-weight counterparts in all the tasks.

~\\
\noindent
\textbf{Notation.} 
Given a probability measure $\mu$ on $\RBB^d$, we denote $\mu \in \PM_2(\RBB^d)$ if its second moment is finite.
For a given functional $\FM(\mu):\PM_2(\RBB^d) \to \RBB$, 
$\frac{\delta \FM(\rho)}{\delta \rho}:\RBB^d \to \RBB$ denote its first variation at $\mu=\rho$.
Besides, we use $\nabla$ and $\nabla \cdot ()$ to denote the gradient and the divergence operator, respectively.

\section{Preliminaries}
\subsection{Particle-Based Variational Inference Methods}
When dealing with Bayesian inference tasks, classical Variational Inference methods approximate the target posterior $\pi$ with an easy-to-sample 
    distribution $\mu$, and recast the inference task as an optimization problem over $\mathcal{P}_2(\mathbb{R}^d)$ (or its subspace) \cite{ranganath2014black}: 
\begin{align}
    \min_{\mu\in \mathcal{P}_2(\mathbb{R}^d)}\mathcal{F}(\mu):=\mathcal{D}(\mu|\pi).
\end{align}

To solve this optimization problem, one can consider a descent flow of $\mathcal{F}(\mu)$ in the Wasserstein space,
which transports any initial distribution $\mu_0$ towards the target distribution $\pi$ \cite{wibisono2018sampling}.
Specifically, the descent flow of $\FM(\mu)$ is described by the following \emph{continuity equation} \cite{ambrosio2008gradient,santambrogio2017euclidean}: 
\begin{align}\label{pde}
    \partial_t\mu_t = -  \nabla\cdot(\mu_t\vb_{\mu_t}) , 
\end{align}
where $\vb_{\mu_t} : \mathbb{R}^d\to\mathbb{R}^d$ is a vector field that defines the direction of position transportation.
To ensure a descent of $\mathcal{F}(\mu_t)$ over time $t$, the vector field $\vb_{\mu_t}$ should satisfy the following inequality 
    \cite{ambrosio2008gradient}: 
\begin{align}\label{dissipation_w}
    \frac{\mathrm{d}\mathcal{F}(\mu_t)}{\mathrm{d}t} = \int \langle\nabla \frac{\delta \mathcal{F}(\mu_t)}{\delta \mu_t},\vb_{\mu_t} \rangle \dr \mu_t \leq 0 . 
\end{align} 
A straightforward choice of $\vb_{\mu_t}$ is $\vb_{\mu_t} = -\nabla\frac{\delta \mathcal{F}(\mu_t)}{\delta \mu_t}$, 
    which is actually the steepest descent direction of $\FM(\mu_t)$. 
From now on, we fix this choice of $\vb_{\mu_t}$ in \eqref{pde}.
We note that the continuity equation \eqref{pde} with this $\vb_{\mu_t}$ is also known as the \emph{Wasserstein gradient flow} of $\mathcal{F}$.

To simulate the Wasserstein gradient flow of $\mathcal{F}$, 
existing ParVIs evolve a set of $M$ fixed-weight particles 
and use the empirical distribution $\tilde{\mu}_t = \sum_{i=1}^M a^i_t \delta_{\xb^i_t}$ to approximate $\mu_t$ in \eqref{pde},
where $\xb_t^i$ and $a^i_t$ (usually set to $1/M$) denote the position and the weight of the $i$-th particle at time $t$, respectively.
Specifically, ParVIs update the position of  each particle $\xb^i_t$ according to the following finite-particle \emph{position  transport  approximation} of 
    the continuity equation \eqref{pde} \cite{chen2018unified, liu2017stein,craig2016blob}: 
\begin{align}\label{ode}
    \mathrm{d}\xb^i_t = \vb_{\tilde{\mu}_t}(\xb^i_t)\mathrm{d} t, 
\end{align}
where $\vb_{\tilde{\mu}_t}$ is an approximation of $\vb_{\mu_t}$ through the empirical distribution $\tilde{\mu}_t$. 
It can be verified that the empirical distribution $\tilde \mu_t$ weakly converges to $\mu_t$ defined in \eqref{pde} when $M \to \infty$ under mild conditions 
    \cite{korba2020non, liu2017stein,liu2018stein}. 
Therefore, one can obtain different ParVIs by choosing proper $\vb_{\tilde{\mu}_t}$ and discretizing \eqref{ode} with certain scheme 
    (the first-order explicit Euler discretization is set as a default).
Note that adjusting weight $a^i_t$ according to the continuity equation \eqref{pde} is difficult, 
    as directly discretizing \eqref{pde} needs to calculate the second-order derivative of the first variation due to the divergence operator. 
Hence, existing ParVIs only update the position $\xb^i_t$ and keep the weight $a^i_t$ fixed during the whole procedure. 

To develop a ParVI method, it remains to select a proper dissimilarity functional $\FM$ and construct an approximation $\vb_{\tilde \mu_t}$ 
    of the vector field $\vb_{\mu_t}$.
From the seminal SVGD to the subsequent Blob and GFSD, KL-divergence is widely adopted as the underlying dissimilarity functional 
    \cite{liu2016stein,chen2017particle,chen2018unified,liu2019understanding,liu2019understanding_mcmc, zhang2020stochastic}.
The associated vector field is defined as follows \cite{jordan1998variational, liu2019understanding}: 
\begin{align*}
    \vb_{\mu_t} \!=\! -\nabla\frac{\delta \text{KL}(\mu_t|\pi)}{\delta \mu_t}  \!=\!-\!\nabla \log \frac{\mu_t}{\pi}\!=\! \nabla\log{\pi} \!-\! \nabla\log{\mu_t}. \nonumber
\end{align*}
As $\nabla \log \mu_t$ is undefined with the empirical distribution $\tilde{\mu}_t$, the KL-divergence based ParVIs use different approaches to construct suitable approximations to the vector field.
In SVGD, \citeauthor{liu2016stein}~\shortcite{liu2016stein} restrict the vector field $\vb_{\mu_t}$ within the unit ball of a Reproducing Kernel Hilbert Space (RKHS), 
    and propose to approximate $\vb_{\mu_t}$ by  
\begin{align}\label{svgd}
    \vb_{\tilde{\mu}_t}(\cdot) = \mathbb{E}_{\xb'\sim\tilde{\mu}_t}\left[K(\xb', \cdot)\log{\pi(\xb)} + \nabla_{\xb'}K(\xb', \cdot)\right], 
\end{align}
where $K$ is a kernel function, such as the Radial Basis Function (RBF) kernel.
Subsequently, Blob \cite{chen2018unified} reformulates the intractable term $\nabla\log{\mu_t}$ by partly smoothing the density with a kernel function $K$: $\nabla\left(\frac{\delta}{\delta\tilde{\mu}_t}\mathbb{E}_{\tilde{\mu}_t}\left[\tilde{\mu}_t*K\right]\right)$, 
    while GFSD \cite{liu2019understanding} directly approximates $\mu_t$ by $\tilde{\mu}_t * K$, where $*$ denotes the convolution operator.
Recently, Kernel Stein Discrepancy Descent (KSDD) \cite{korba2021kernel} method considers the Kernel Stein Discrepancy (KSD) \cite{liu2016kernelized} as the 
    underlying functional $\mathcal{F}$, whose first variation is compatible with empirical distribution and can be calculated directly.


\subsection{Fisher-Rao Distance and Reaction Flow}
The Fisher-Rao distance \cite{rao1945information, kakutani1948equivalence} is a metric defined for general positive Radon measures and allows 
    comparing measures with mass variations.
For two probability measures $(\beta_0, \beta_1) \in \PM_2(\RBB^d)$, if they have densities $(\rho_0, \rho_1)$ w.r.t. the Lebesgue measure, 
their Fisher-Rao distance is defined as follows
    \cite{nikulin2001hellinger}:
    \begin{align*}
        \text{FR}(\beta_0, \beta_1) = \|\sqrt{\rho_0} - \sqrt{\rho_1}\|_{L^2}.
    \end{align*}
The positive Radon measure space equipped with the Fisher-Rao distance is known as the Fisher-Rao space.
For a given dissimilarity functional $\FM(\mu)$, its gradient flow in the Fisher-Rao space, also named as the Fisher-Rao reaction flow 
    \cite{gallouet2019unbalanced, gallouet2017jko}, is described by the following equation 
    \cite{wang2019accelerated, liero2016optimal}: 
    \begin{align}\label{reaction_flow}
        \partial_t \mu_t = -\alpha_{\mu_t} \mu_t, \ 
        \alpha_{\mu_t} = \frac{\delta\mathcal{F}(\mu_t)}{\delta \mu_t} - \int{\frac{\delta\mathcal{F}(\mu_t)}{\delta \mu_t}}\mathrm{d}\mu_t,  
    \end{align}
    where $\alpha_{\mu_t}:\mathbb{R}^d\to\mathbb{R}$ represents a construction/destruction function of mass. 
Since the average variation of $\mu_t$ equals zero due to the extra integral term $\int{\frac{\delta\mathcal{F}(\mu_t)}{\delta \mu_t}}\mathrm{d}\mu_t$, 
the total mass of $\mu_t$ is conserved during the whole procedure \cite{lu2019accelerating, rotskoff2019global}.
The target distribution $\pi$ is actually an invariant distribution of this flow, as the first variation of the dissimilarity functional 
    $\mathcal{F}$ vanishes at $\pi$, i.e., $\frac{\delta \mathcal{F}(\pi)}{\delta \pi}= 0$.
It can be verified that with a proper $\FM$, the process \eqref{reaction_flow} starting from a given $\mu_0$ evolves towards the target distribution $\pi$ 
    \cite{kondratyev2016new}.

For a fixed position $\xb$, the process \eqref{reaction_flow} provides an effective way to adjust its density (weight) $\mu_t(\xb)$ at each time $t$ 
    according to the function $\alpha_{\mu_t}$.
Thus, given a set of weighted particles and its empirical distribution $\tilde \mu_t$, one can adjust the weight by discretizing the reaction flow with an empirical approximate construction/deconstruction function $\alpha_{\tilde \mu_t}$.
Though the reaction flow has been adopted in particle-based methods in other literature, such as MCMC \cite{lu2019accelerating},  
    global minimization \cite{rotskoff2019global} and generative models \cite{mroueh2020unbalanced}, 
    to the best of our knowledge, it has never been adopted in the ParVI literature to adjust the weights of particles.
\section{Methodology}
In this section, we first construct a composite flow that evolves positions and weights of particles simultaneously and investigate its mean-field property.
Then, we develop our DPVI framework by discretizing this composite flow.
We finally provide three effective DPVI algorithms by using different dissimilarity functionals $\mathcal{F}$ (KL-divergence and KSD) and finite-particle approximations.
Besides, we also derive the duplicate/kill variants of our proposed algorithms.

\subsection{Continuous-Time Composite Flow}
Based on the position transport approximation \eqref{ode} for displacing the position  and the Fisher-Rao reaction flow \eqref{reaction_flow} for adjusting weight in previous sections,
we consider the following composite flow that evolves the positions $\xb^i$'s and the weights $a^i$'s of $M$ particles simultaneously.
\begin{align}\label{cmp}
    \left\{
	\begin{aligned}
		\mathrm{d}\xb^i_t &=   \vb_{\tilde \mu_t} \mathrm{d}t, \\
		\mathrm{d} a^i_t &=\textstyle -\left(U_{\tilde{\mu}_t}(\xb^i_t) - \sum_{i=1}^M a^i_t U_{\tilde{\mu}_t}(\xb^i_t)\right)a^i_t\mathrm{d}t,\\
		\tilde{\mu}_t   &= \textstyle\sum_{i=1}^M a^i_t\delta_{\xb^i_t}.
	\end{aligned}\right.
\end{align}
For ease of notation, we use $U_{\mu}$ and $ \vb_{ \mu}$ to denote the first variation of $\mathcal{F}(\mu)$ at $\mu$ and the vector field associated with it, respectively, i.e., 
    $U_{\mu} = \frac{\delta\mathcal{F}(\mu)}{\delta\mu}$ and $\vb_{\mu} = -\nabla U_{\mu}$.
Although the mean-field limit of the empirical distribution with either the position update part $\dr \xb^i_t$ or the weight adjustment part $\dr a^i_t$ alone
    has the target distribution $\pi$ as its stationary distribution, 
    the behaviour of the empirical distribution $\tilde{\mu}_t$ in \eqref{cmp} remains unknown.
Thus, we first investigate the mean-field limit of $\tilde{\mu}_t$ and show that it actually follows the gradient flow of $\FM$ in the Wasserstein-Fisher-Rao 
    space when $M\to\infty$.
Due to limited space, we refer readers to the Appendix for detailed proof of all the propositions.
\begin{proposition}\label{proposition_1}
    Suppose the empirical distribution $\mut^M_0$ of $M$ weighted particles weakly converges to a distribution $\mu_0$ when $M \to \infty$. 
    Then, the path of \eqref{cmp} starting from $\mut^M_0$ 
    weakly converges to a solution of the following partial differential equation starting from $\mu_0$ as $M\to\infty$:
    \begin{align}\label{gf_wfr}
        \partial_t \mu_t \!=\! \nabla\cdot\left(\mu_t\nabla U_{\mu_t}\right) \!-\! \left(U_{\mu_t} \!-\! \int{U_{\mu_t}\mathrm{d}\mu_t}\right)\mu_t,
    \end{align}
    which is actually the gradient flow of $\FM$ in the Wasserstein-Fisher-Rao space.
\end{proposition}
Compared to the Wasserstein gradient flow \eqref{pde} of $\FM$, the Wasserstein-Fisher-Rao gradient flow \eqref{gf_wfr} has an additional density adjustment part 
    (second term in the r.h.s. of equation \eqref{gf_wfr}).
It can be verified that \eqref{gf_wfr} results in a faster decrease of $\FM$ than the Wasserstein gradient flow used in 
    the fixed-weight ParVIs due to the additional density adjustment part.
\begin{proposition}\label{proposition_2}
Consider the Wasserstein-Fisher-Rao gradient flow \eqref{gf_wfr} of a given functional $\FM$, the differentiation of $\FM(\mu_t)$ with respect to the time $t$ 
    satisfies: 
    \begin{align}\label{wfr-descent}
        \frac{\mathrm{d}\mathcal{F}(\mu_t)}{\mathrm{d}t} = &-\int{\left\|\nabla\left(\frac{\delta\mathcal{F}(\mu_t)}{\delta\mu_t}\right) \right\|^2\mathrm{d}\mu_t}\\
        &-\left(\int{\left|\frac{\delta\mathcal{F}(\mu_t)}{\delta\mu_t}\right|^2\mathrm{d}\mu_t} - \left(\int{\frac{\delta\mathcal{F}(\mu_t)}{\delta\mu_t}\mathrm{d}\mu_t}\right)^2\right).\nonumber
    \end{align}
\end{proposition}
Note that the first part on the r.h.s. of \eqref{wfr-descent} comes from the position update $\dr \xb_i^t$ and 
    the second part is resulted from the weight adjustment $\dr a_i^t$.
As a result, it can be verified that the mean-field limit of the composite flow leads to a faster decrease of $\mathcal{F}$ compared with either the Wasserstein gradient flow or the Fisher-Rao reaction flow of $\FM$.
With proper dissimilarity functional $\FM$ such as the KL-divergence, it can be further proved that the Wasserstein-Fisher-Rao gradient flow \eqref{gf_wfr} 
    evolves to the target $\pi$ faster than both the Wasserstein gradient flow \eqref{pde} and the Fisher-Rao reaction flow under mild conditions \cite{lu2019accelerating}.
For a general dissimilarity $\FM$, it remains an open problem that whether the extra density adjustment would result in a faster convergence to the target $\pi$, 
    and we leave it as an interesting future work.

\subsection{Dynamic-Weight ParVI Framework}\label{sec:dparvi}
Generally, it is impossible to obtain an analytic solution of the continuous composite flow \eqref{cmp}, thus a numerical integration method is required to 
    derive an approximate solution.
Note that any numerical solver, such as the implicit Euler method \cite{platen2010numerical} and higher-order Runge-Kutta method \cite{butcher1964implicit} can be used. 
Here, we adopt the first-order explicit Euler discretization \cite{suli2003introduction} since it is simple and easy-to-implement,
    and propose our Dynamic-weight Particle-based Variational Inference (DPVI) framework, as listed in Algorithm \ref{alg:algorithm}.

Starting from $M$ weighted particles located at $\{\xb_0^i\}_{i=1}^M$ with weights $\{a_0^i\}_{i=0}^M$, DPVI first 
    updates the positions of particles according to the following rule: 
\begin{align}\label{discrete_position}
    \xb^i_{k+1} = \xb^i_{k} + \eta \vb_{\tilde{\mu}_k}(\xb^i_k), 
\end{align}
where $\tilde{\mu}_k = \sum_{i=1}^M a^i_k\delta_{\xb^i_{k}}$.
Then, it adjusts the particles' weights following 
\begin{align}\label{dynamic_weight}
	a^i_{k+1} = a^i_k - \lambda\eta\bar{U}_{\tilde{\mu}_{k+1/2}}(\xb^i_{k+1})a^i_{k}, 
\end{align}
where $\bar{U}_{\tilde{\mu}_{k+1/2}} = U_{\tilde{\mu}_{k+1/2}} - \sum_{i=1}^M a^i_k U_{\tilde{\mu}_{k+1/2}}(\xb^i_{k+1})$, and 
$\tilde{\mu}_{k+1/2} = \sum_{i=1}^M a^i_k\delta_{\xb^i_{k+1}}$ represents the empirical distribution after the position update \eqref{discrete_position}.
Here, we assume that a suitable empirical approximation $U_{\tilde{\mu}_{k}}$ of the first variation is already constructed, and we will discuss this
    comprehensively in the following subsection.
It can be verified that the total mass of $\tilde{\mu}_k$ is conserved as the following equality holds: 
    \begin{align*}
        \textstyle\sum_{i=1}^M \eta \bar{U}_{\tilde{\mu}_{k+1/2}}(\xb^i_{k+1})a^i_k  = 0,
    \end{align*}
Thus, $\tilde{\mu}_k$ remains a valid probability distribution during the whole procedure of DPVI, i.e. $\sum_i a^i_k = 1$ for all $k$.
    
In developing our DPVI framework, we adopt a Gauss-Siedel-type strategy to update the position $\xb^i_k$ and the weight $a^i_{k}$, i.e., we adjust the 
    weight based on the newly obtained position $\xb_{k+1}^i$ in the $k$-th iteration.
As the weight adjustment step \eqref{dynamic_weight} only involves calculating the first variation approximation $U_{\tilde \mu_{k+1/2}}$, it would bring little extra computational 
    cost compared with the position update step \eqref{discrete_position}, which involves calculating the gradient of $U_{\tilde \mu_{k}}$.
Besides, one can further reduce the computational cost by adopting the Jacobi-type strategy, i.e., update weight $a_{k}^i$ in \eqref{dynamic_weight} with $U_{\tilde \mu_{k}}$ at position $\xb_{k}^i$. 
In this case, the weight adjustment step boils down to $M$ scalar additions since the term $U_{\tilde \mu_{k}}$ can be directly obtained when calculating 
    $\vb_{\tilde{\mu}_k} = \nabla U_{\tilde \mu_{k}}$ in the position update step \eqref{discrete_position}.
Here, we adopt the Gauss-Siedel-type update strategy as it usually has better empirical performance without bringing much additional computational cost compared with the 
    Jacobi-type update strategy.

\begin{algorithm}[tb]
\caption{Dynamic-weight Particle-based Variational Inference (DPVI) Framework}
\label{alg:algorithm}
\textbf{Input}: 
Initial distribution $\tilde{\mu}_0 = \sum_{i=1}^M a^i_0 \delta_{\xb^i_0}$, step-size $\eta$, weight parameter $\lambda$.	
\begin{algorithmic}[1] 
\FOR{$k = 0,1,...,T-1$}
\FOR{$i = 1,2,...,M$}
\STATE Update positions $\xb^i_{k+1}$'s according to \eqref{discrete_position}.
\ENDFOR 
\FOR{$i = 1,2,...,M$}
\STATE  Adjust weights $a^i_{k+1}$'s according to \eqref{dynamic_weight}. 
\ENDFOR 
\ENDFOR
\STATE \textbf{Output}:$\tilde{\mu}_T = \sum_{i=1}^M a^i_T \delta_{\xb^i_T}$.
\end{algorithmic}
\end{algorithm}

\subsection{DPVI algorithms and their duplicate/kill variants}\label{sec:dparvi-alg}
To derive an efficient DPVI algorithm, it remains to decide the underlying functional $\mathcal{F}$, and construct proper empirical approximations to its first variation $U_{\mu}$ 
    and the associated vector field $ \vb_{\mu} = -\nabla U_{\mu}$ (denoted as $U_{\tilde{\mu}}$ and $\vb_{\tilde{\mu}}$, respectively).
Unfortunately, there exists no systematic approach to design proper approximations $U_{\tilde{\mu}}$ and $\vb_{\tilde{\mu}}$ for arbitrary dissimilarity functional $\FM$.
Here, we propose three efficient DPVI algorithms based on different approximations utilized in existing fixed-weight ParVIs,
two with the KL-divergence as the underlying functional $\FM$ and one with the KSD.

\paragraph{KL-divergence as $\mathcal{F}$.} 
As we have discussed in the preliminaries, a large portion of existing fixed-weight ParVIs adopt the KL-divergence as underlying functional $\FM$, 
    whose associated vector field and first variation are defined as follows: 
\begin{align*}
    \begin{array}{ll}
    	\vb_{\mu}(\xb)= \nabla\log{\pi(\xb)} - \nabla\log{\mu(\xb)}, 
    	\\
    	U_{\mu}(\xb)  = \log{\mu(\xb)} - \log{\pi(\xb)}.
    \end{array}
\end{align*}
In order to deal with the intractable $\log{\mu(\xb)}$ which is undefined with empirical distribution $\tilde{\mu}_k$, 
we adopt the approximation techniques used in the fixed-weight ParVI methods GFSD \cite{liu2019understanding} and Blob \cite{chen2018unified}, 
and derive two KL-divergence based DPVI algorithms. 

\emph{D-GFSD-CA algorithm.} 
Our first DPVI algorithm utilizes the same approximation technique in the fixed-weight GFSD algorithm, and we name it as D-GFSD-CA algorithm (``CA'' for \emph{Continuous Adjustment}). 
Specifically, we directly approximate $\mu$ by smoothing the empirical distribution $\tilde{\mu}$ with a kernel function $K$: 
    $\hat{\mu} = \tilde{\mu}*K  = \sum_{i=1}^Ma^iK(\cdot, \xb^i)$, which leads to the following approximations: 
    \begin{align}
        \vb_{\tilde{\mu}_k}(\xb)  = &\nabla\log{\pi(\xb)} - \frac{\sum_{i=1}^M a^i_k\nabla_{\xb}K(\xb, \xb^i_k)}{\sum_{i=1}^M a^i_k K(\xb,\xb^i_k)}, \label{gfsd-v}
        \\
        U_{\tilde{\mu}_{k+1/2}}(\xb)  \!=\! & \!-\! \log{\pi(\xb)}\! +\! \textstyle\log{\sum_{i=1}^M a^i_k K(\xb,\xb^i_{k+1})}.\label{gfsd-u}
    \end{align}

\emph{D-Blob-CA algorithm.}
We name our second DPVI algorithm as D-Blob-CA, which uses the same approximation technique adopted by the fixed-weight Blob method.
In particular, we reformulate the intractable term $\log{\mu}$ from the perspective of its first variation $\frac{\delta}{\delta \mu}\mathbb{E}_{\mu}\left[\log{\mu}\right]$ and 
    partly smooth the density with a kernel function $K$:
    \begin{align*}
        &\frac{\delta}{\delta \tilde{\mu}}\mathbb{E}_{\tilde{\mu}}\left[\log{(\tilde{\mu}*K)}\right]\nonumber =
        \log{(\tilde{\mu}*K)} + \frac{\tilde{\mu}}{\tilde{\mu}*K}*K \\
        = & \textstyle\log{\sum_{i=1}^Ma^iK(\cdot,\xb^i)} + \displaystyle\sum_{i=1}^M\frac{a^iK(\cdot, \xb^i)}{\sum_{l=1}^Ma^lK(\xb^i, \xb^l)}.\nonumber
    \end{align*}
This leads to the following approximation results: 
    \begin{align}
        \vb_{\tilde{\mu}_k}(\xb)  = &\nabla\log{\pi(\xb)} - \frac{\sum_{i=1}^M a^i_k\nabla_{\xb}K(\xb, \xb^i_k)}{\sum_{i=1}^M a^i_k K(\xb,\xb^i_k)} \nonumber\\
            & - \sum_{i=1}^M{\frac{a^i_k\nabla_{\xb}K(\xb,\xb^i_k)}{\sum_{l=1}^Ma^l_kK(\xb^i_k,\xb^l_k)}},\label{blob-v} \\
        U_{\tilde{\mu}_{k+1/2}} (\xb) = & - \log{\pi(\xb)} + \textstyle\log{\sum_{i=1}^M a^i_k K(\xb,\xb^i_{k+1})} \nonumber\\
            & + \sum_{i=1}^M{\frac{a^i_kK(\xb,\xb^i_{k+1})}{\sum_{l=1}^Ma^l_kK(\xb^i_{k+1},\xb^l_{k+1})}}.\label{blob-u}
    \end{align}

\begin{remark}\label{blob-better}
    In the above approximations, we call the terms defined through the interaction with other particles
    as the repulsive terms.
    It can be observed that the Blob-type approximations \eqref{blob-v} and \eqref{blob-u} have an extra repulsive term (the term in the second line)
    compared to the GFSD-type approximations \eqref{gfsd-v} and \eqref{gfsd-u}.
    Practically, this extra repulsive term would drive the particles away from each other further,
    and result in a better exploration of particles in the probability space.
    Actually, the Blob-type methods usually outperforms the GFSD-type methods empirically. 
\end{remark}

Since GFSD and Blob (partly) smooth the original empirical distribution $\tilde{\mu}$ with a kernel function $K$, the underlying evolutionary 
    distribution is actually a smoothed version of $\tilde{\mu}$. 
To update the positions and the weights in the smoothed empirical distribution, one should solve a system of linear 
    equations to obtain the new positions $\xb^i_{k+1}$'s and weights $a^i_{k+1}$'s in the $k$-th iteration.
Nevertheless, with a proper kernel function $K$, such as the RBF kernel, the density $\mu(\xb^i)$ at a given position $\xb^i$
    mainly comes from its corresponding weight $a^i$.
Hence, we can still update the positions and weights according to \eqref{discrete_position} and \eqref{dynamic_weight}, respectively.

\paragraph{KSD as $\mathcal{F}$.}
Except for the KL-divergence, KSD is recently adopted as the dissimilarity functional in the fixed-weight ParVI method KSDD \cite{korba2021kernel}, 
    whose first variation and the corresponding vector field are defined as
    \begin{align}\label{ksdd_vu}
        \begin{array}{ll}
        \vb_{\mu}(\xb) =  \mathbb{E}_{\xb'\sim\mu}\left[\nabla_{\xb}k_{\pi}(\xb',\xb)\right], \\
        U_{\mu}(\xb) =  \mathbb{E}_{\xb'\sim\mu}\left[k_{\pi}(\xb',\xb)\right].
        \end{array}
    \end{align}
    Here, $k_{\pi}$ denotes the Stein kernel \cite{liu2016kernelized}, and it is defined by the score of $\pi$: $s(\xb) = \nabla\log{\pi(\xb)}$ and a positive 
        semi-definite kernel function $K$:
    \begin{align*}
        k_\pi(\xb,\yb) = & s(\xb)^Ts(\yb)K(\xb,\yb) + s(\xb)^T\nabla_\yb K(\xb, \yb)\\ & + \nabla_\xb K(\xb,\yb)^T s(\yb) + 
            \nabla_\xb \cdot\nabla_\yb K(\xb,\yb).
    \end{align*}

\emph{D-KSDD-CA algorithm.}
The vector field $\vb_{\mu}$ and the first variation $U_{\mu}$ in \eqref{ksdd_vu} can be directly approximated via the empirical distribution $\tilde{\mu}$.
We construct the following finite-particle approximations $\vb_{\tilde{\mu}_k}$ and $U_{\tilde{\mu}_{k+1/2}}$: 
    \begin{align*}
        \vb_{\tilde{\mu}_k}(\xb) =& \textstyle\sum_{i=1}^M a^i_k \nabla_\xb k_\pi(\xb^i_k, \xb), \\
        U_{\tilde{\mu}_{k+1/2}}(\xb) = & \textstyle\sum_{i=1}^M a^i_k  k_\pi(\xb^i_{k+1}, \xb).
    \end{align*}
By using with these approximations in DPVI, we obtain the D-KSDD-CA algorithm.

\paragraph{The duplicate/kill variants.}
In fact, there is a probabilistic discretization strategy of the approximate reaction flow in \eqref{cmp}, 
    which has been used by particle-based methods in other literature 
    \cite{lu2019accelerating,rotskoff2019global}.
This strategy duplicates/kills particle $\xb^i_{k+1}$ according to an exponential clock with instantaneous rate: 
    \begin{align}\label{dk_particles}
        R^i_{k+1} = -\lambda\eta\bar{U}_{\tilde{\mu}_{k+1/2}}(\xb^i_{k+1}).
    \end{align}
Specifically, if $R^i_{k+1}>0$, duplicate the particle $\xb^i_{k+1}$ with probability $1-\exp{(-R^i_{k+1})}$, and kill another one with 
    uniform probability to conserve the total mass; if $R^i_{k+1}<0$, kill the particle $\xb^i_{k+1}$ with probability $1 - \exp{(R^i_{k+1})}$, 
    and duplicate another one with uniform probability.

This Duplicate/Kill (DK) strategy could also be used as an alternative to the CA strategy \eqref{dynamic_weight} in the DPVI framework. 
By replacing CA with DK in D-GFSD-CA, D-Blob-CA and D-KSDD-CA, 
    we derive three DPVI variants, denoted as D-GFSD-DK, D-Blob-DK and D-KSDD-DK, respectively. 
Note that due to the deterministic position update fashion of ParVI algorithms, the particles may collapse to few positions during this duplicate/kill procedure.
Thus, we inject an additional Gaussian noise with variance scaled by step-size for the duplicated particle to avoid particle collapse.

    The DK strategy has also been used in a dynamics based MCMC method,
        Birth Death Langevin Sampling (BDLS) \cite{lu2019accelerating}.
    Specifically, BDLS update $M$ particles independently according to the Langevin dynamics, 
        a special form of the Wasserstein gradient flow of the KL-divergence \cite{chen2018unified}, 
        and use the DK strategy to duplicate/kill particles.
    Though both the dynamics based MCMC methods and the ParVIs are suitable for Bayesian inference,
        ParVIs are usually more particle-efficient as they use the repulsive force between particles 
        to drive the particles away from each other and explore the whole space \cite{liu2019understanding}.
    Actually, we also observe that the performance of BDLS is worse than 
        the KL-divergence based DPVI algorithms 
        due to the lack of interaction between particles.
    

\section{Experiments}
In this section, we conduct empirical studies with our DPVI algorithms (D-GFSD-CA, D-Blob-CA, and D-KSDD-CA), 
	their duplicate/kill variants (D-GFSD-DK, D-Blob-DK, and D-KSDD-DK), and the fixed-weight ParVI algorithms (SVGD, GFSD, Blob and KSDD).
Besides, we also include BDLS \cite{lu2019accelerating} and the Unadjusted Langevin Dynamics (ULD) method \cite{ma2015complete} (the fixed-weight counterpart of BDLS) as our baselines.
We compare the performance of these algorithms on two simulations, i.e., bivariate Gaussian model and Gaussian mixture model (GMM),
	and two real-world applications, i.e. Gaussian Process (GP) regression and Bayesian neural network.
For all the algorithms, the particles' weights are initialized to be equal.
We repeat all the experiments 10 times and report the average results.
Due to limited space, only parts of the results are reported in this section.
We refer readers to the Appendix for the experiment on the bivariate Gaussian model and additional results for GMM.

\begin{table}[tb]
	\centering
	\begin{tabular}{c|ccccc}
		\toprule
		\multirow{2}*{Algorithm} & \multicolumn{5}{c}{Number of particles}  \\
		~    & 5 &10 &20 &50 &100\\
		\hline
		ULD      & 2.467 & 1.818 & 1.398 & 0.977 & 0.835 \\
		BDLS 	 & 2.549 & 1.694 & 1.368 & 0.974 & 0.746 \\
		SVGD     & 1.963 & 1.339 & 1.094 & 0.917 & 0.648 \\
		\hline
		GFSD     & 2.044 & 1.489 & 1.292 & 1.080 & 0.902 \\
		D-GFSD-DK& 1.791 & 1.306 & 1.129 & 0.791 & 0.693 \\
		D-GFSD-CA& 1.618 & 1.096 & 0.884 & 0.642 & 0.496 \\
		\hline
		Blob     & 1.970 & 1.390 & 1.150 & 0.888 & 0.596 \\
		D-Blob-DK& 1.736 & 1.193 & 0.881 & 0.639 & 0.530 \\
		D-Blob-CA& \textbf{1.532} & \textbf{1.014} & \textbf{0.763} & \textbf{0.516} & \textbf{0.388} \\
		\hline
		KSDD     & 1.964 & 1.379 & 1.183 & 1.086 & 0.936 \\
		D-KSDD-DK& 2.011 & 1.452 & 1.172 & 1.089 & 0.919 \\
		D-KSDD-CA& 1.948 & 1.366 & 1.174 & 1.054 & 0.932 \\
		\bottomrule
	\end{tabular}
	\caption{$W_2$ distances with different number of particles.}
	\label{gmm_w2}
	\vspace{-5mm}
\end{table}

\subsection{Gaussian Mixture Model}\label{gmm_exp}
We consider approximating a 2-D Gaussian mixture model with two components, weighted by $1/3$ and $2/3$ respectively.
To investigate the influence of particle number $M$ in fixed-weight ParVIs and the dynamic-weight algorithms, 
we run all the algorithms with $M\in\{5, 10, 20, 50, 100\}$.

In Table \ref{gmm_w2}, we report the 2-Wasserstein ($W_2$) distance between the empirical distribution generated by each algorithm and the target distribution. 
We generate 2100 samples from the target distribution $\pi$ as reference to evaluate the $W_2$ distance 
	by using the POT library \cite{flamary2021pot}. 
It can be observed that both the CA and DK weight strategies contribute to obtain a higher approximation accuracy, 
	and CA usually results in better performance as it allows weights to vary in $[0,1]$ continuously.
The DPVI algorithms constantly outperform their fixed-weight counterparts with the same or even less number of particles.
For instance, D-GFSD-CA with $M=20$ achieves a lower $W_2$ than the fixed-weight GFSD algorithm with $M=100$. 
Moreover, the results also show that, with a fixed particle number $M$, D-Blob-CA achieves the best performance among all the algorithms,
	due to both the dynamic weight strategy and extra repulsive term.
Besides, the KL-divergence based DPVIs outperform BDLS due to the particle efficiency of DPVI, 
	as discussed in the last paragraph of Section \ref{sec:dparvi-alg}. 
Note that the KSDD-type algorithms (KSDD, D-KSDD-CA, and D-KSDD-DK) perform poorly in this task.
Actually, this phenomenon has already been reported in KSDD \cite{korba2021kernel}, 
and the authors claim that particles in KSDD may get stuck in spurious local minima when dealing with multi-mode models.

\begin{figure}[tb]\centering
	\subfigure{\includegraphics[width=.33\linewidth]{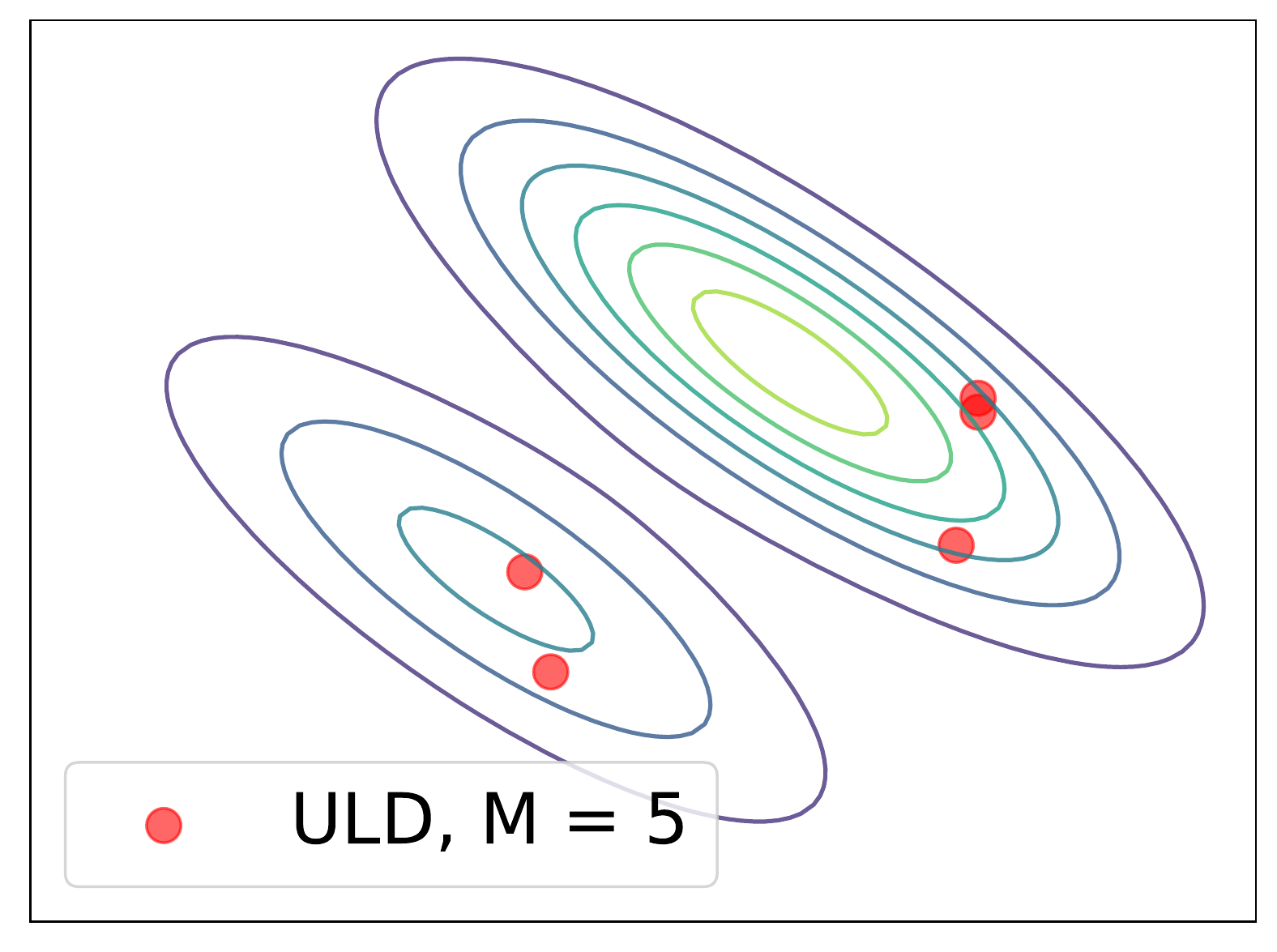}}\hfill
	\subfigure{\includegraphics[width=.33\linewidth]{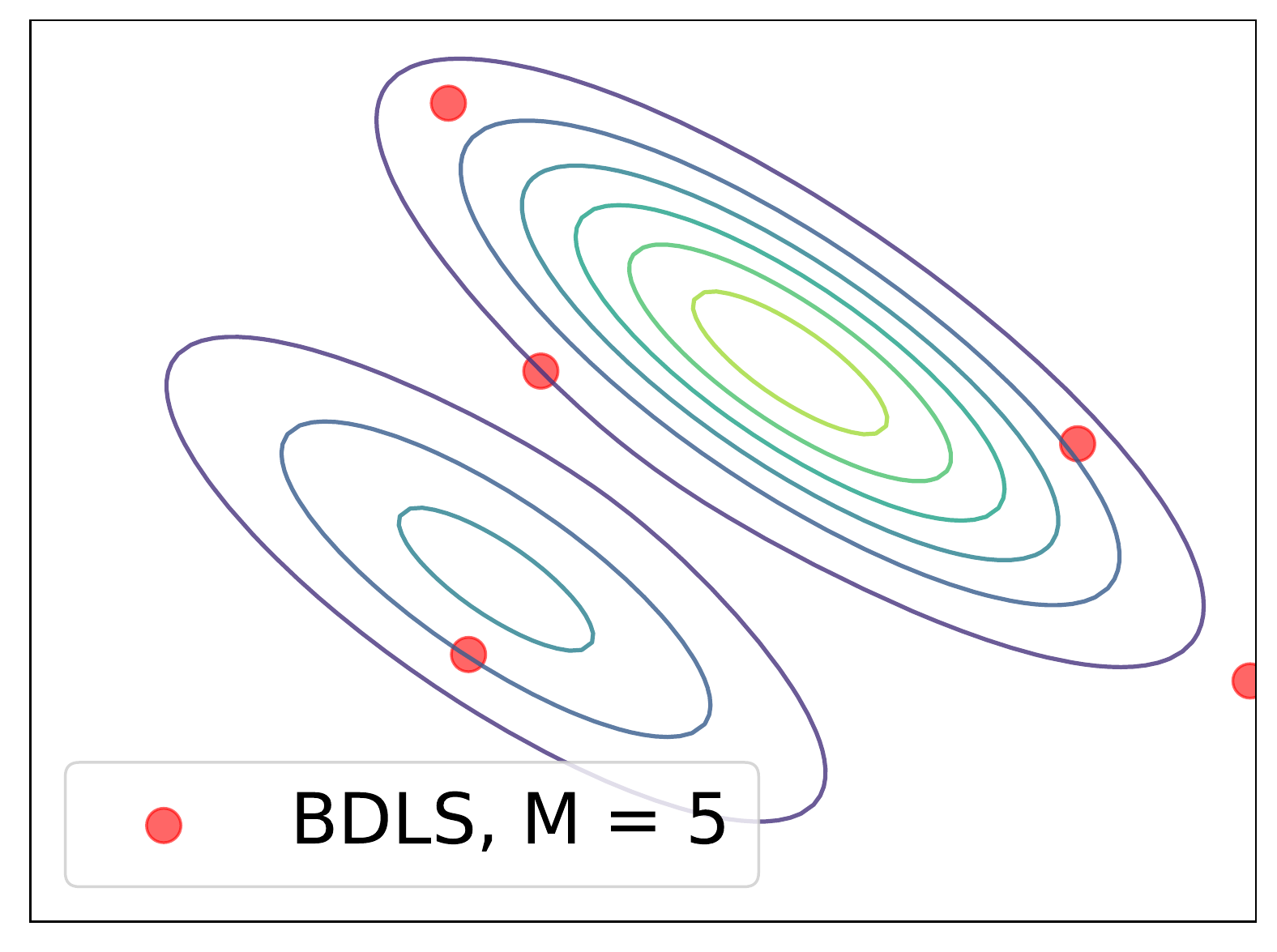}}\hfill
	\subfigure{\includegraphics[width=.33\linewidth]{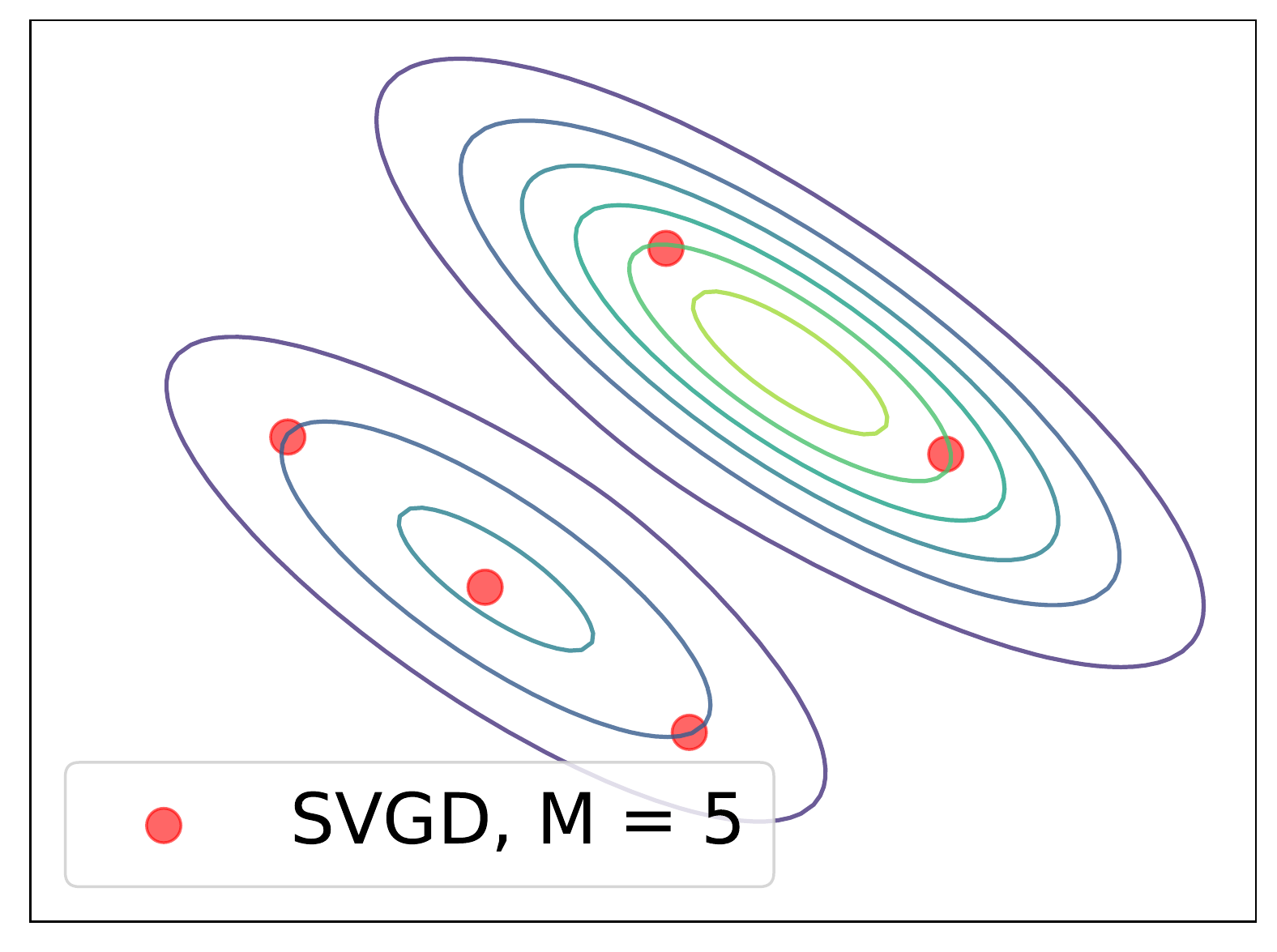}}
	\quad
	
	\vspace{-4mm}
	\subfigure{\includegraphics[width=.33\linewidth]{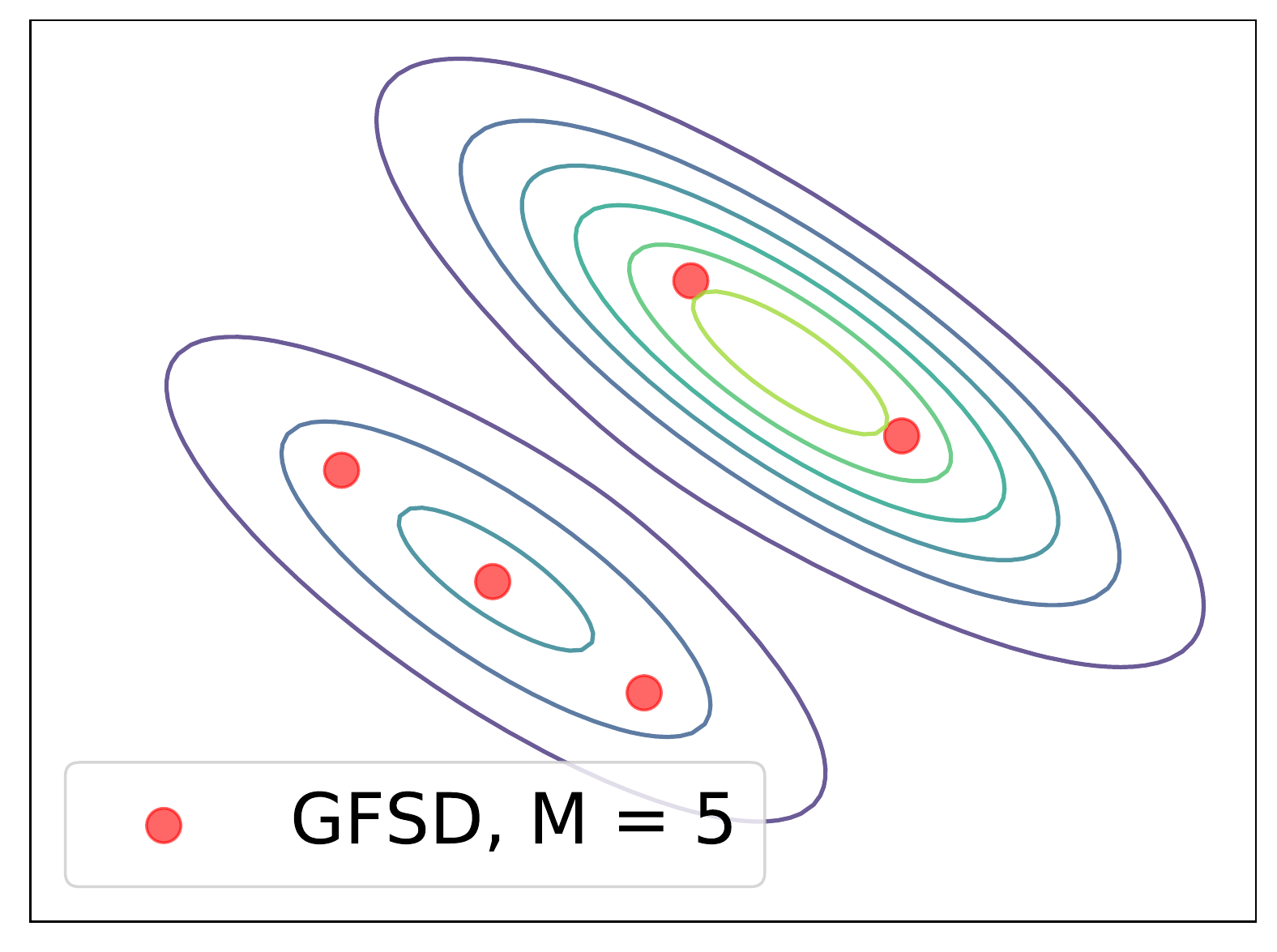}}\hfill
	\subfigure{\includegraphics[width=.33\linewidth]{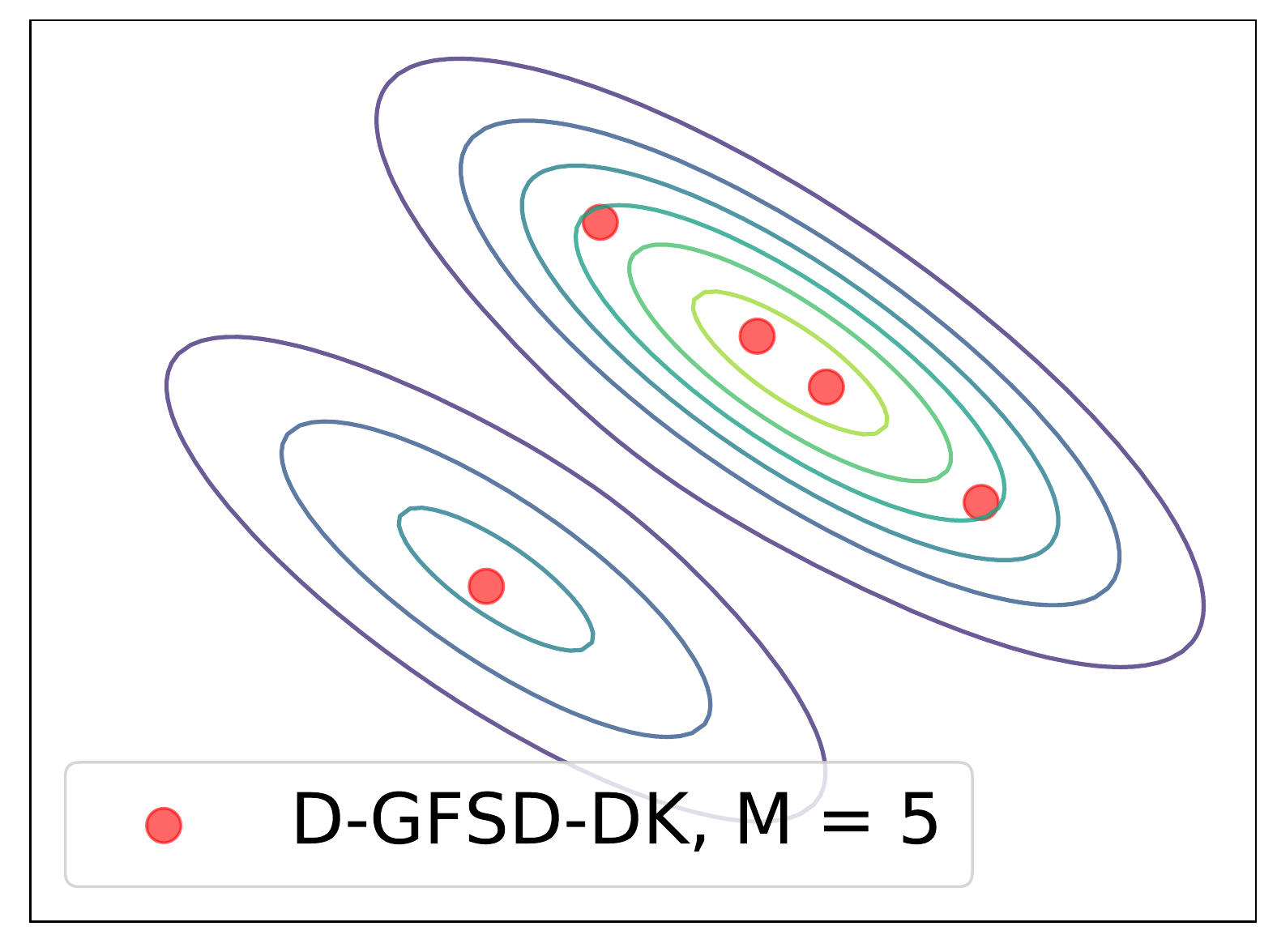}}\hfill
	\subfigure{\includegraphics[width=.33\linewidth]{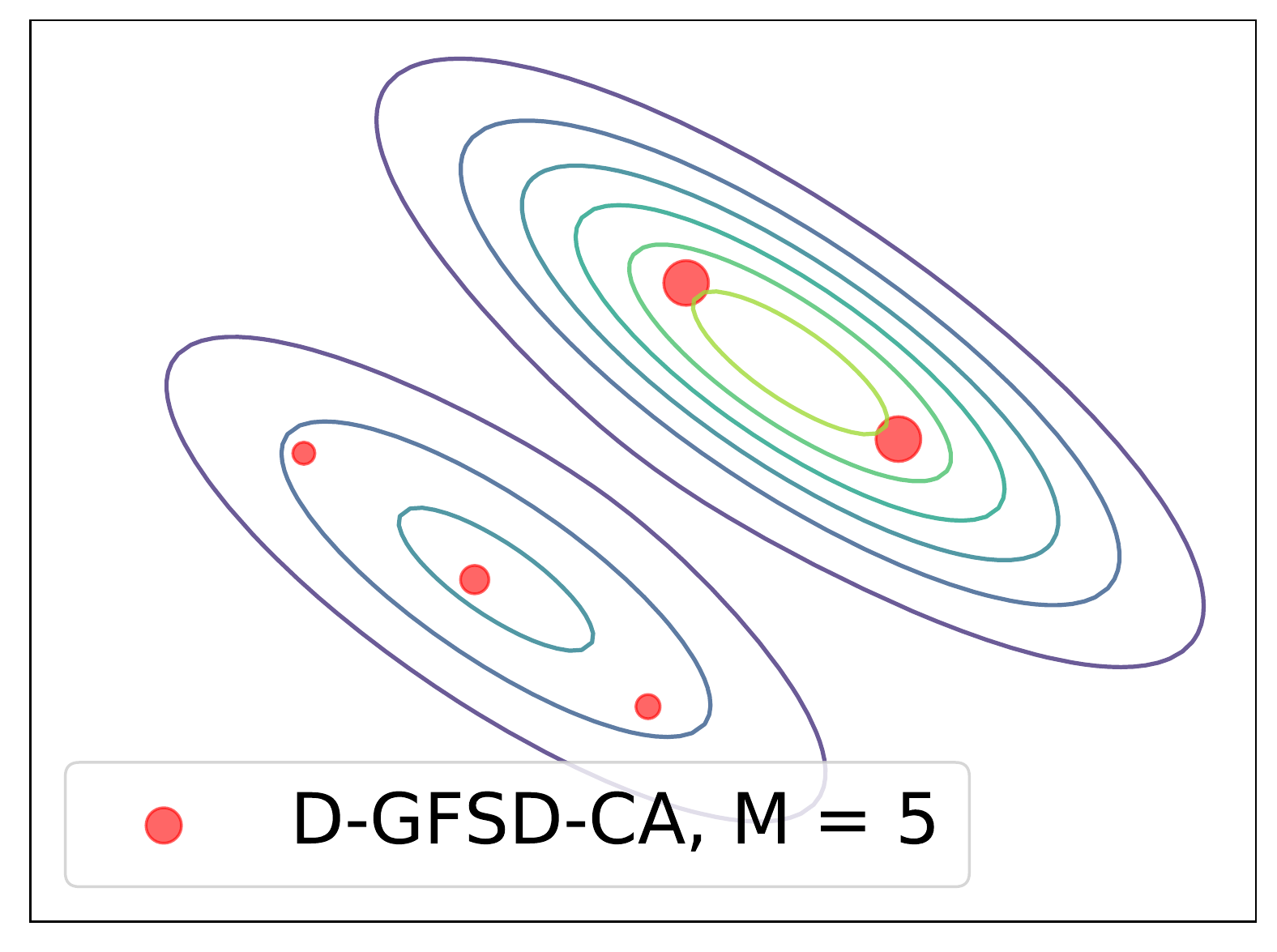}}
	\quad 
	
	\vspace{-4mm}
	\subfigure{\includegraphics[width=.33\linewidth]{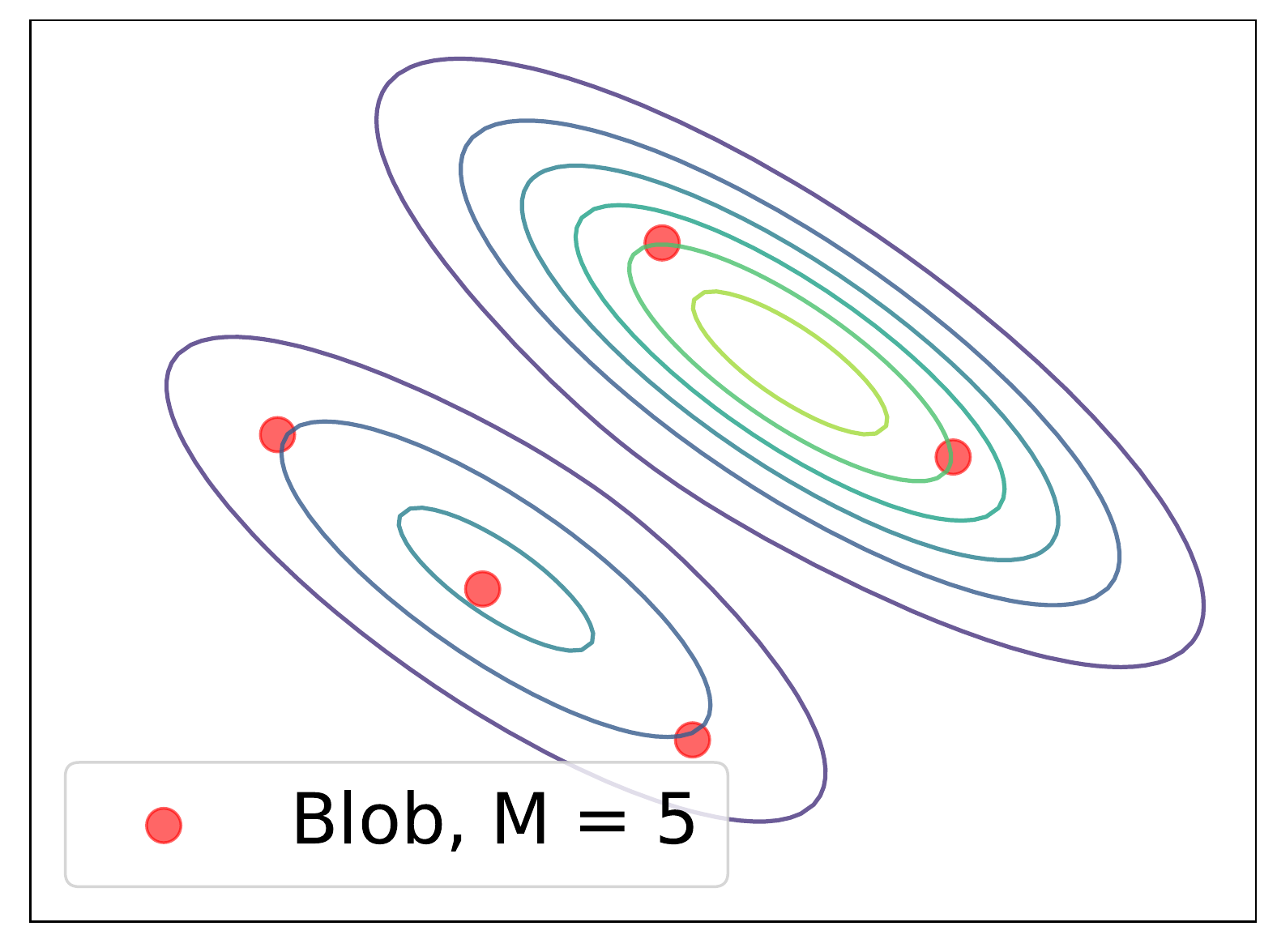}}\hfill
	\subfigure{\includegraphics[width=.33\linewidth]{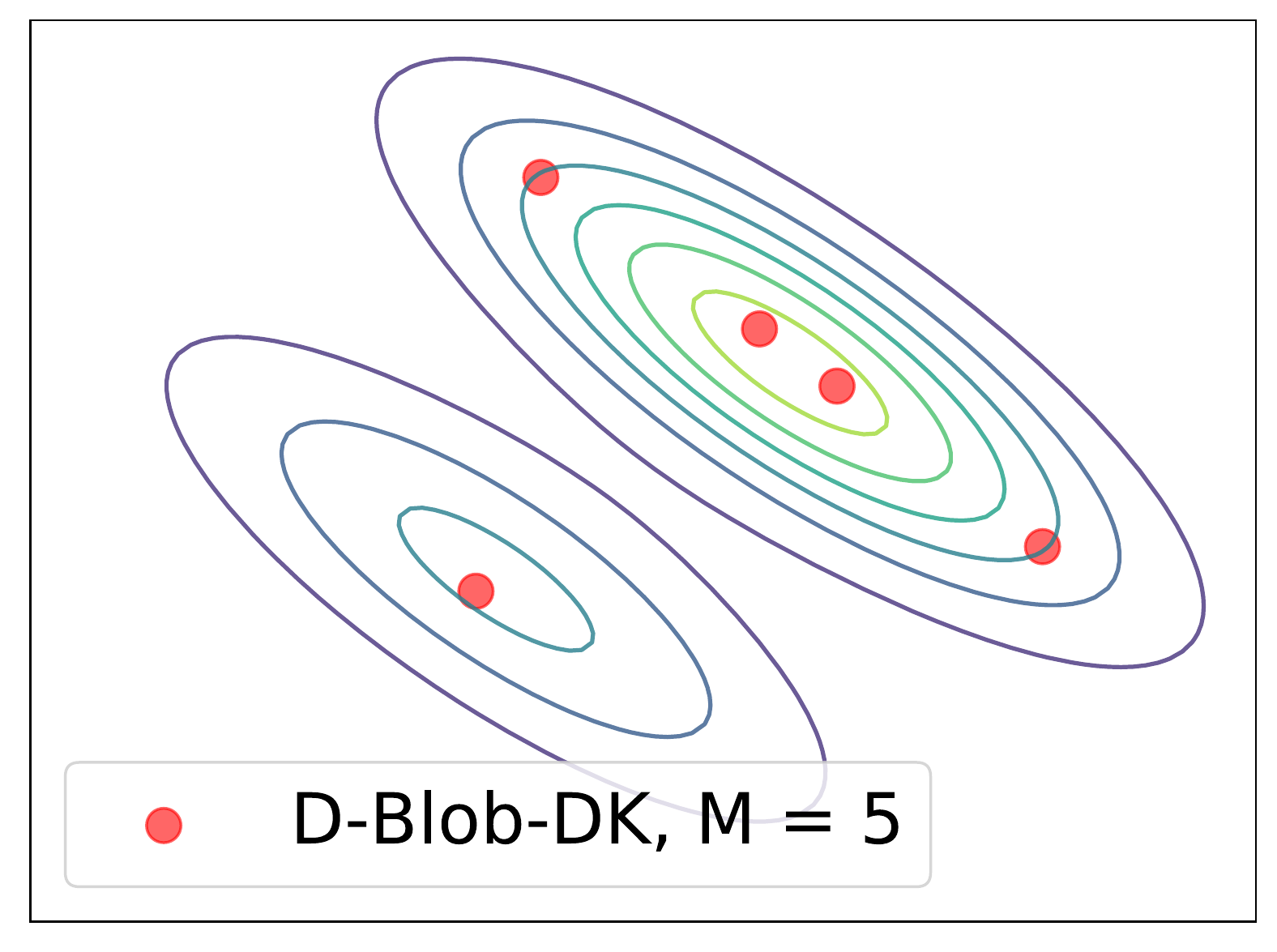}}\hfill
	\subfigure{\includegraphics[width=.33\linewidth]{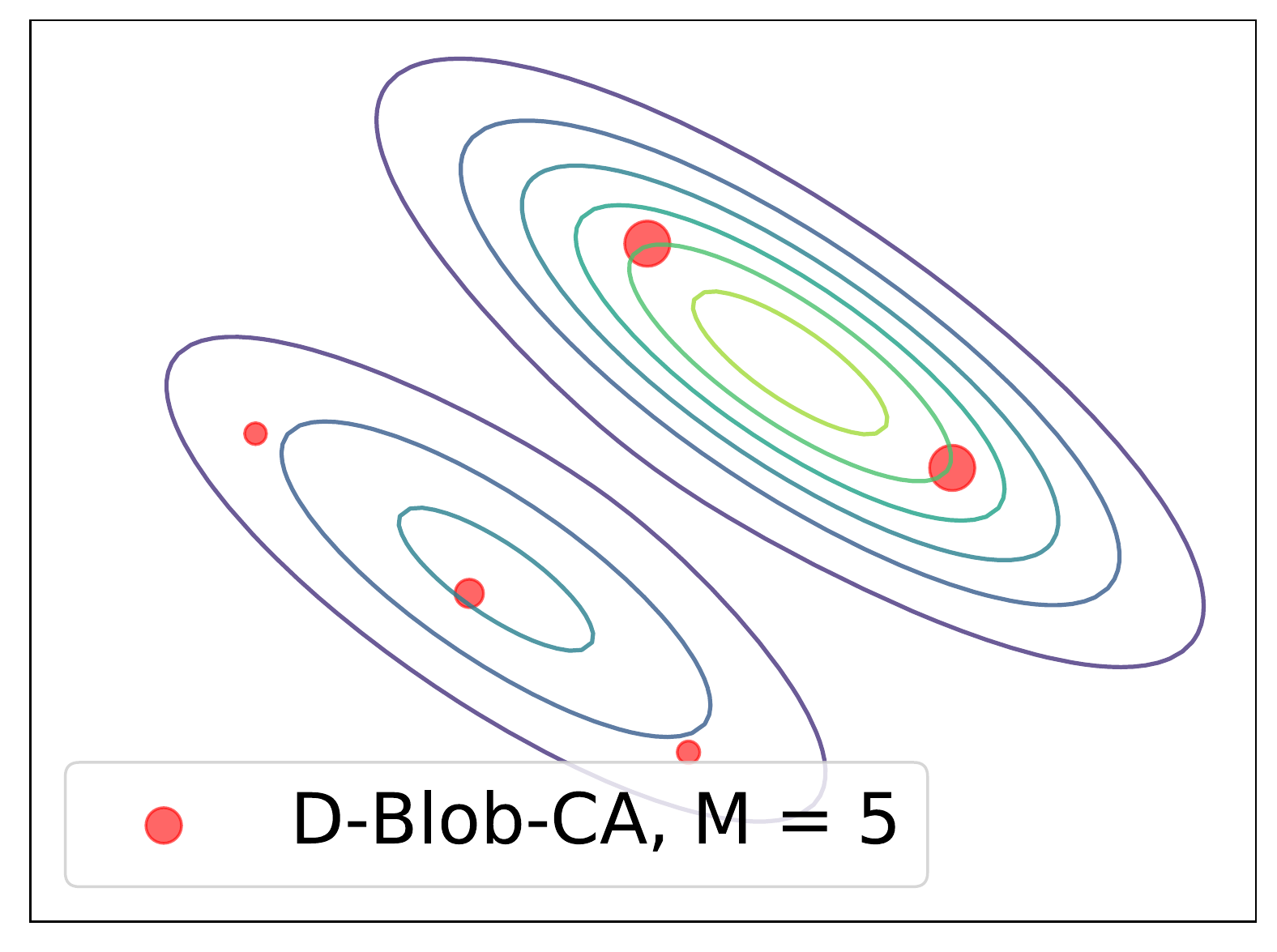}}
	\quad 
	
	\vspace{-4mm}
	\subfigure{\includegraphics[width=.33\linewidth]{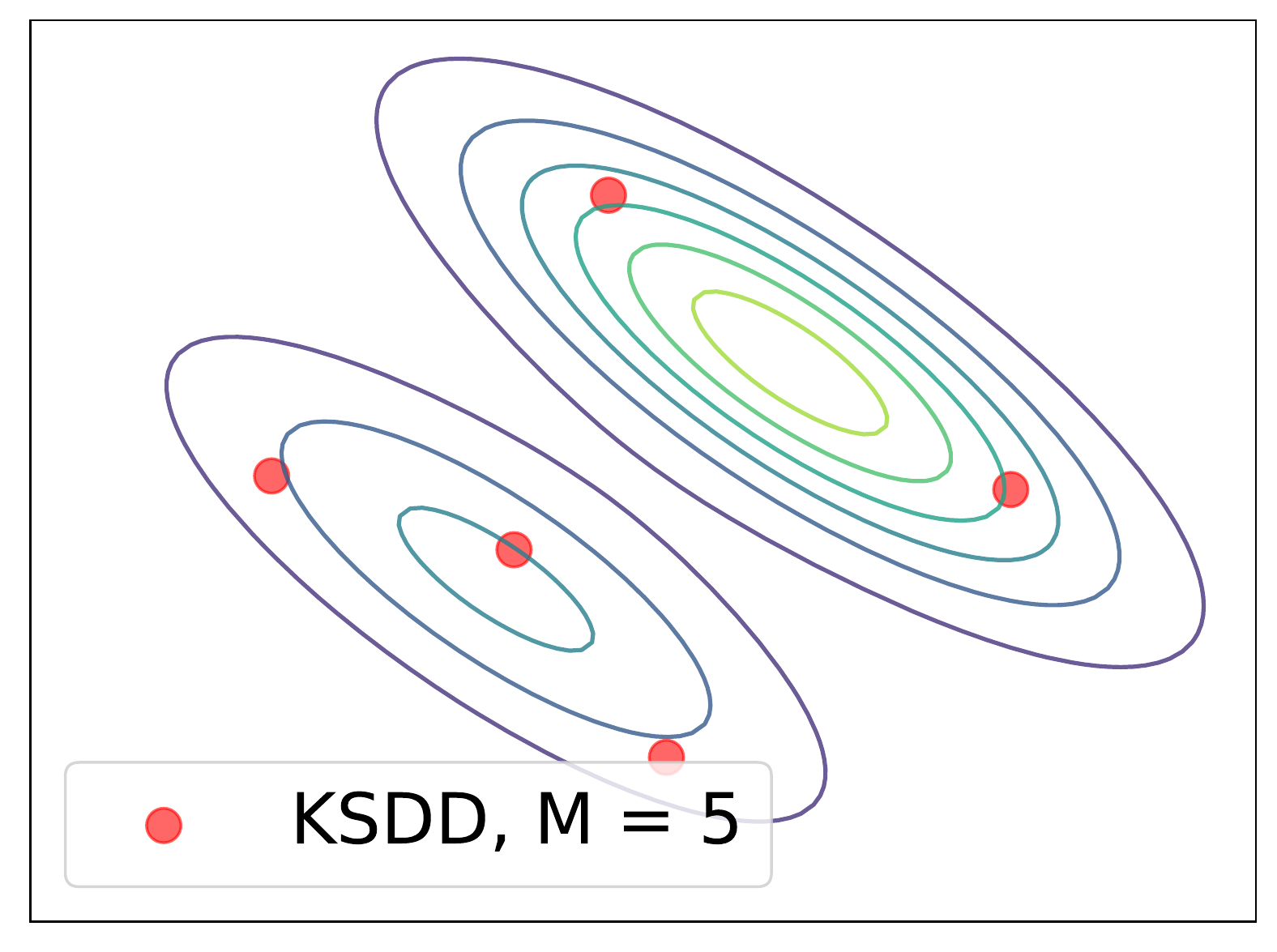}}\hfill
	\subfigure{\includegraphics[width=.33\linewidth]{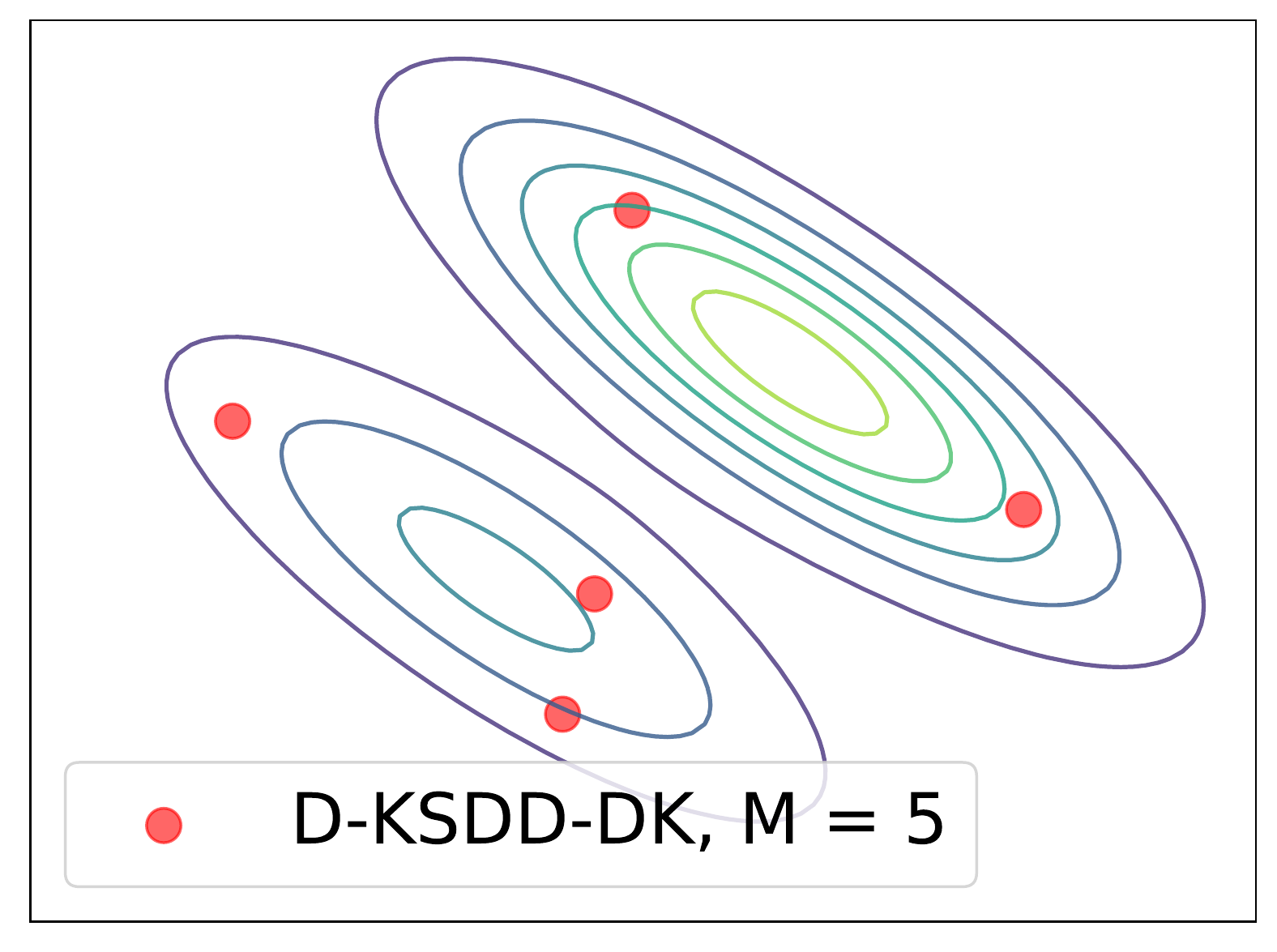}}\hfill
	\subfigure{\includegraphics[width=.33\linewidth]{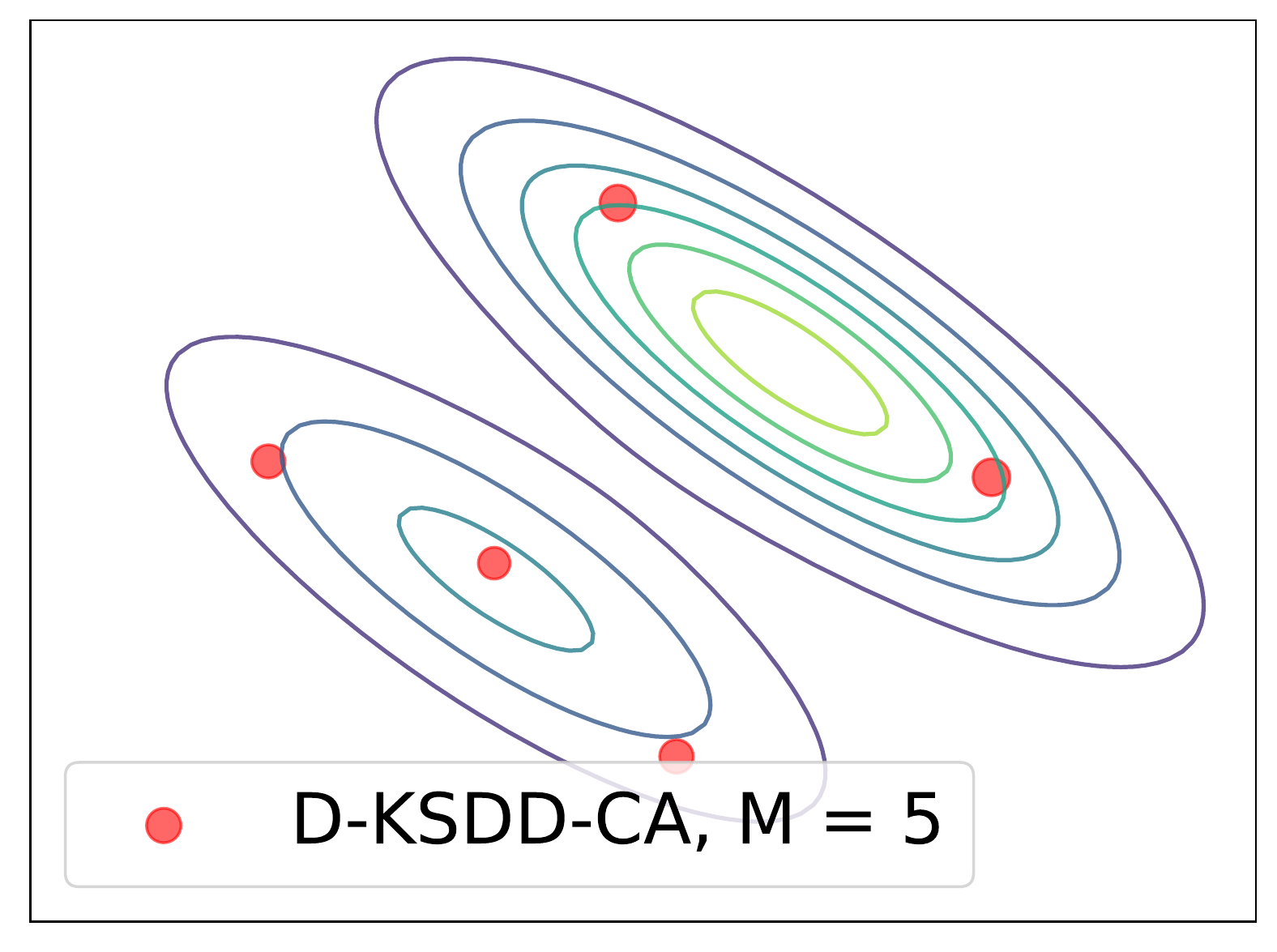}}
	\vspace{-4mm}
	\caption{\label{samples_gmm} Results on approximating a Gaussian mixture distribution with 5 particles.}
	\vspace{-5mm}
\end{figure}

In Figure \ref{samples_gmm}, we plot the output of each algorithm when $M=5$, 
	and the particle size is proportional to its weight. 
From this figure, we can observe that, compared with the fixed-weight ParVI algorithms,
	both the CA and DK strategies
	result in more mass of the empirical distribution on the larger component of the target distribution.
Specifically, the DK strategy concentrates more particles on the larger component, 
	due to its ability to duplicate particles with larger target probability densities;
the	CA strategy increases the weight of each particle on the larger component,
	though algorithms with CA strategy put less particles on this component.
Actually, according to the results in the first column of Table \ref{gmm_w2},
  the DPVI algorithms with CA obtain higher approximation accuracy than their DK variants. 

\subsection{Gaussian Process Regression}
The Gaussian Process (GP) model is widely adopted for the uncertainty quantification in regression problems \cite{rasmussen2003gaussian}. 
We follow the experiment setting in \cite{chen2018stein}, and use the dataset LIDAR (denoted as $\mathcal{D} = \{(x_i,y_i)\}^N_{i=1}$) which consists of 221 observations 
of scalar variable $x_i$ and $y_i$.
Denote $\xb = [x_1, x_2,...,x_N]^T$ and $\yb = [y_1, y_2, ..., y_{N}]^T$, the target log-posterior w.r.t. the model parameter 
$\phi = (\phi_1, \phi_2)$ is defined as follows: 
\begin{align*}
	\log{p(\phi|\mathcal{D})}\! = \!-\frac{\yb^T \Kb^{-1}_y \yb}{2} \!-\! \frac{\log{\det{(\Kb_y)}}}{2} \!-\! \log{(1\!+\!\xb^T\xb)}.
\end{align*}
Here, $\Kb_y$ is a covariance function $\Kb_y = \Kb + 0.04\mathbf{I}$ with $\Kb_{i,j} = \exp(\phi_1)\exp(-\exp(\phi_2)(x_i - x_j)^2)$ and 
	$\mathbf{I}$ represents the identity matrix. 
In this task, we set the particle number to $M=128$ for all the algorithms.

\begin{figure}[tb]\centering
	\subfigure{\includegraphics[width=.33\linewidth]{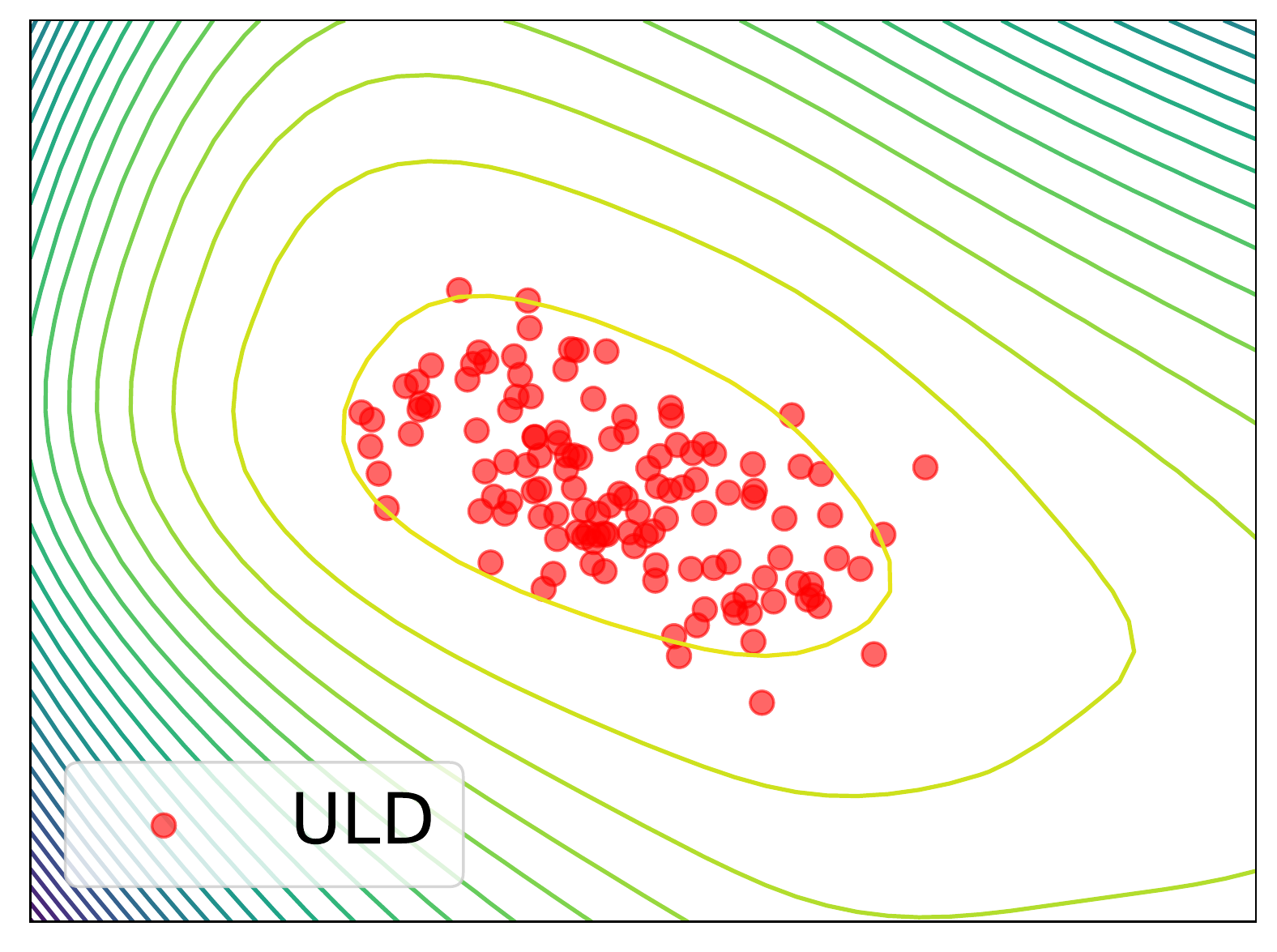}}\hfill
	\subfigure{\includegraphics[width=.33\linewidth]{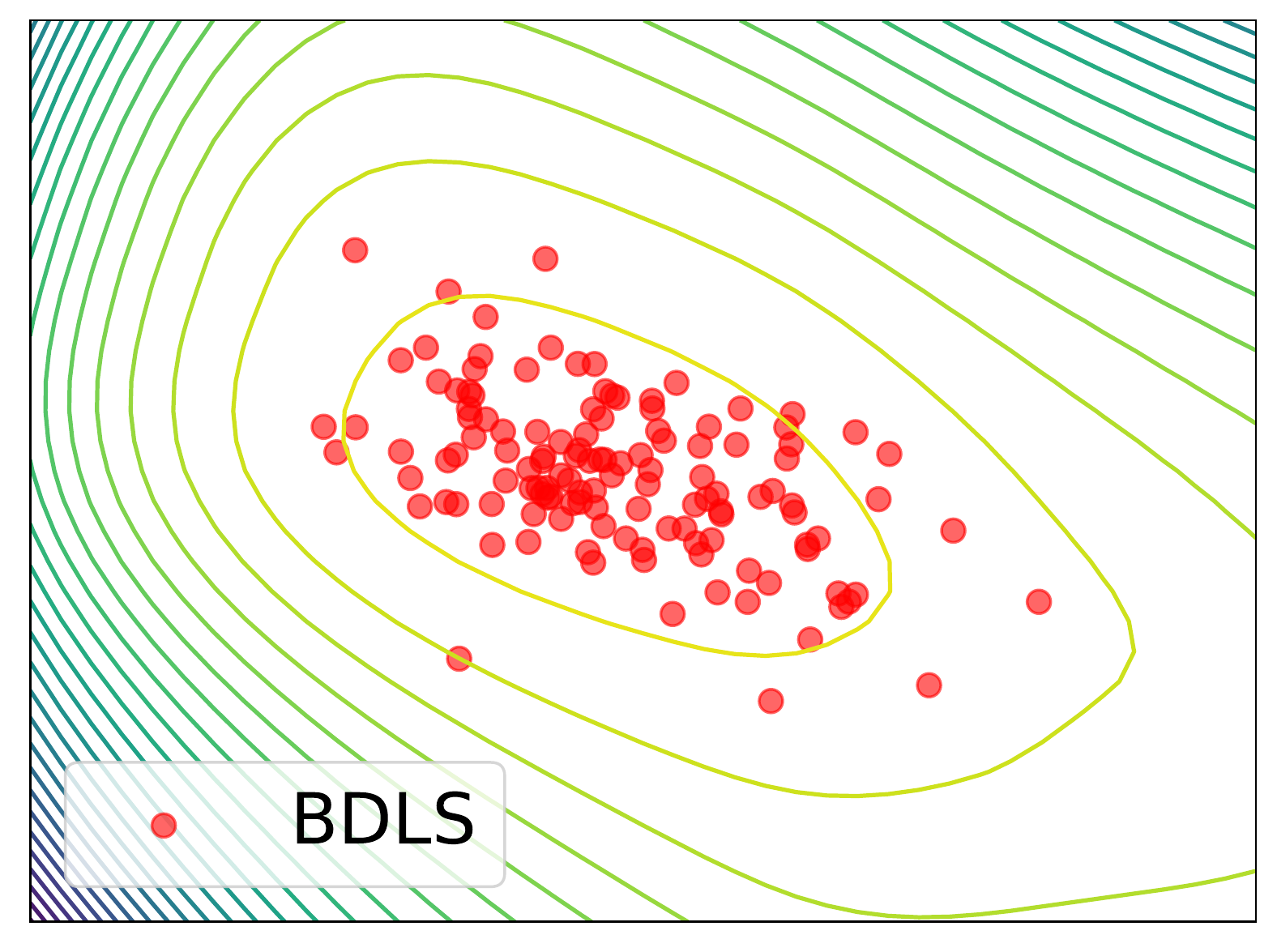}}\hfill
	\subfigure{\includegraphics[width=.33\linewidth]{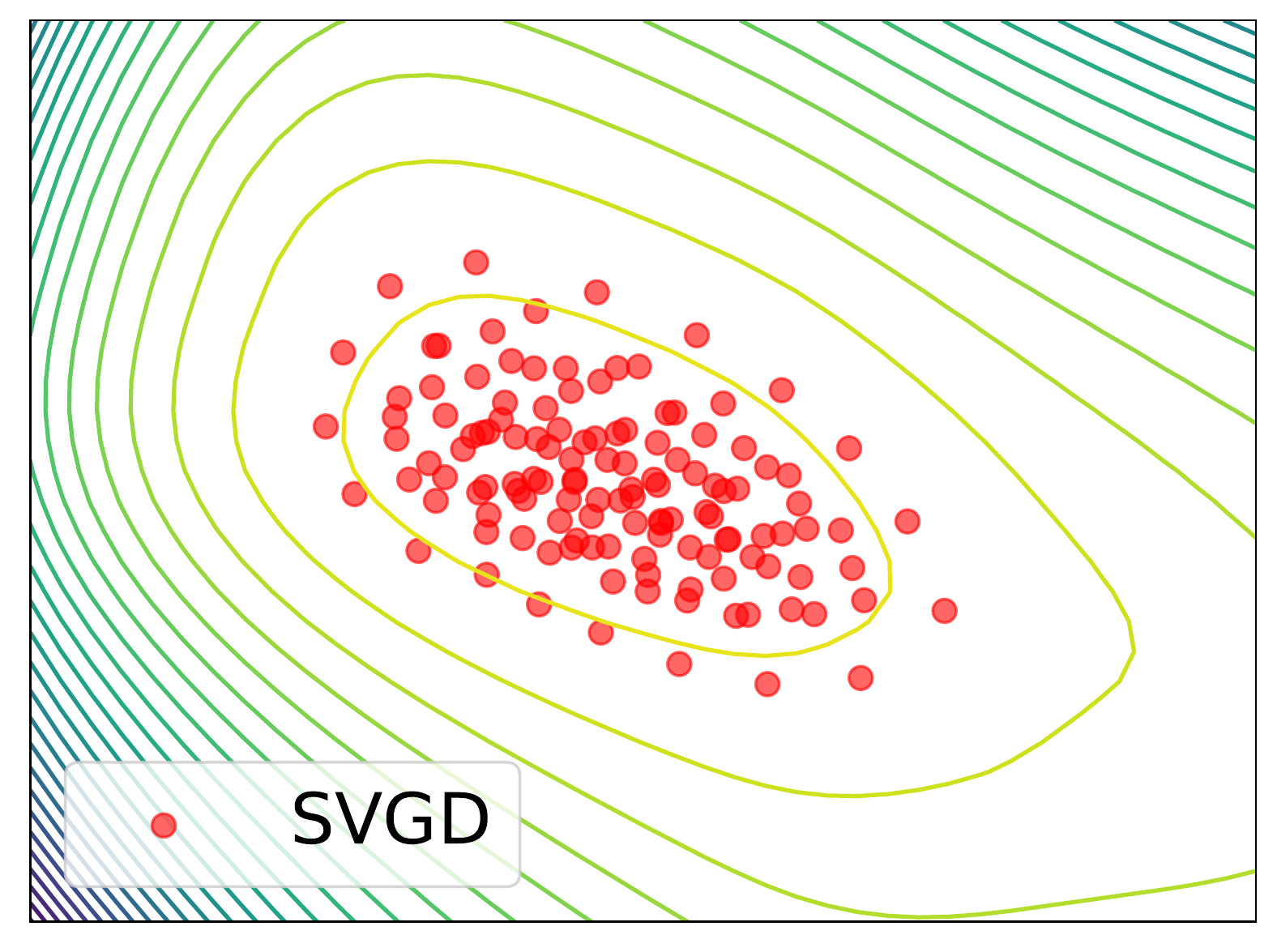}}
	\quad
	
	\vspace{-4mm}
	\subfigure{\includegraphics[width=.33\columnwidth]{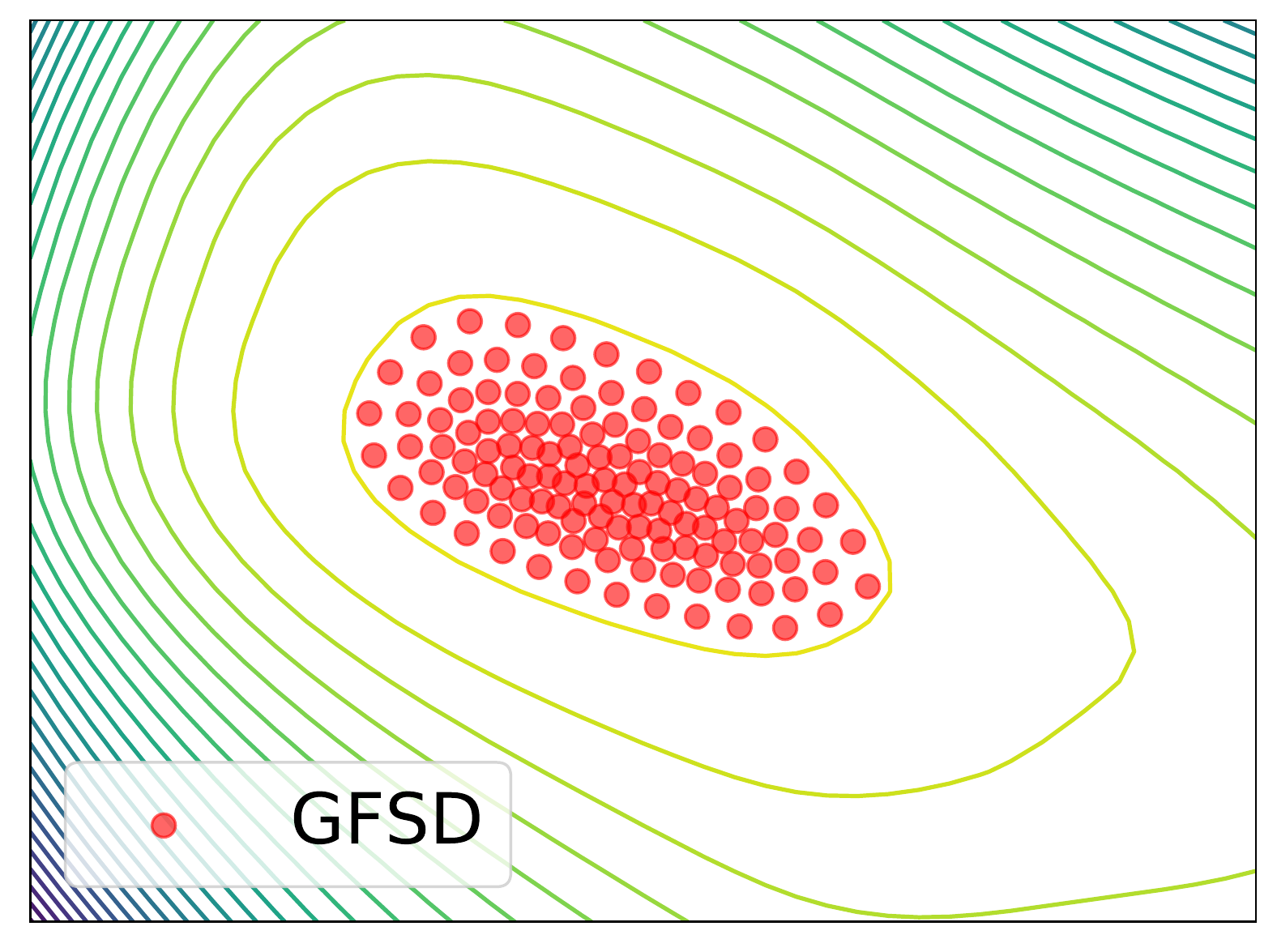}}\hfill
	\subfigure{\includegraphics[width=.33\columnwidth]{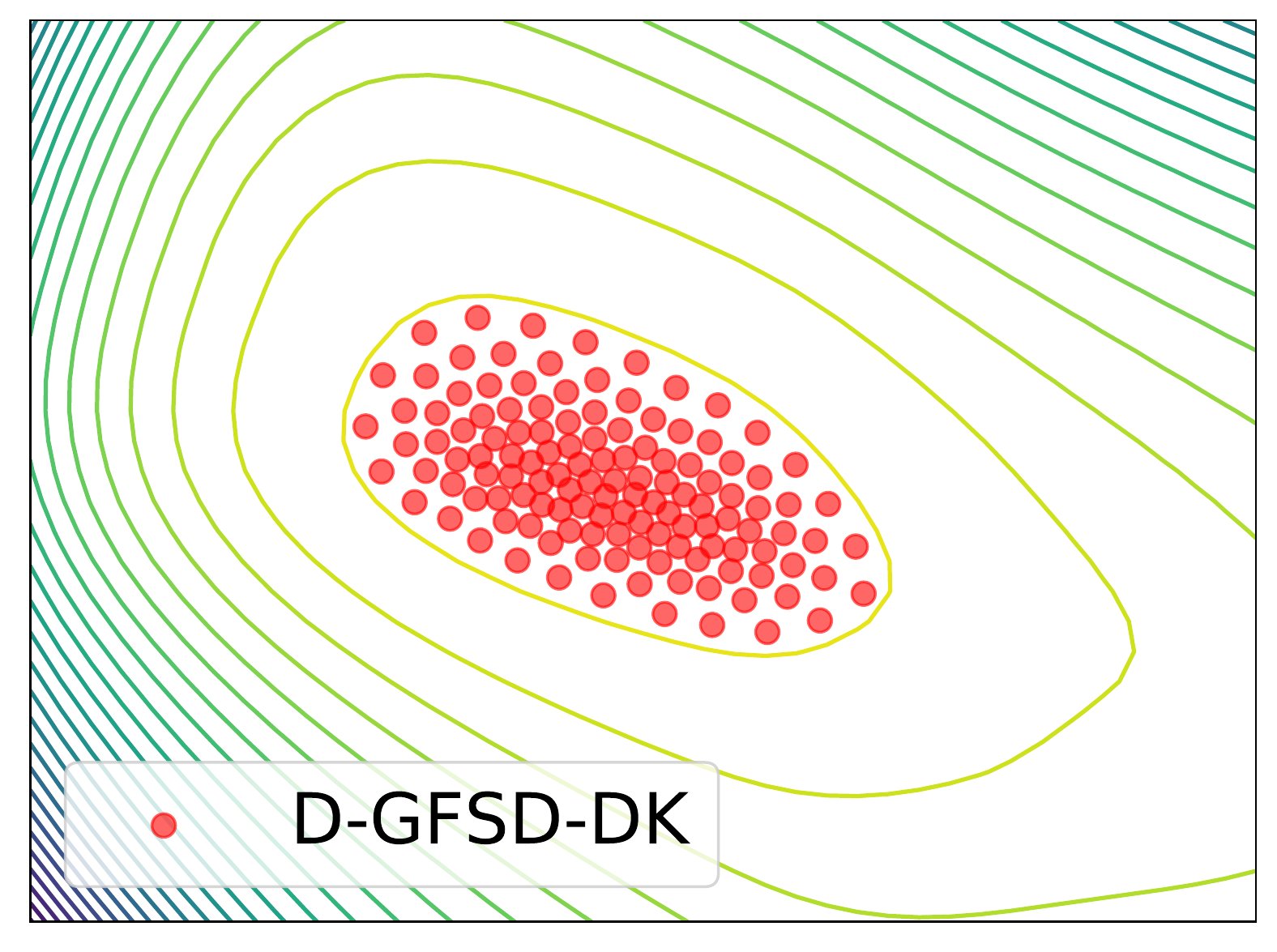}}\hfill
	\subfigure{\includegraphics[width=.33\columnwidth]{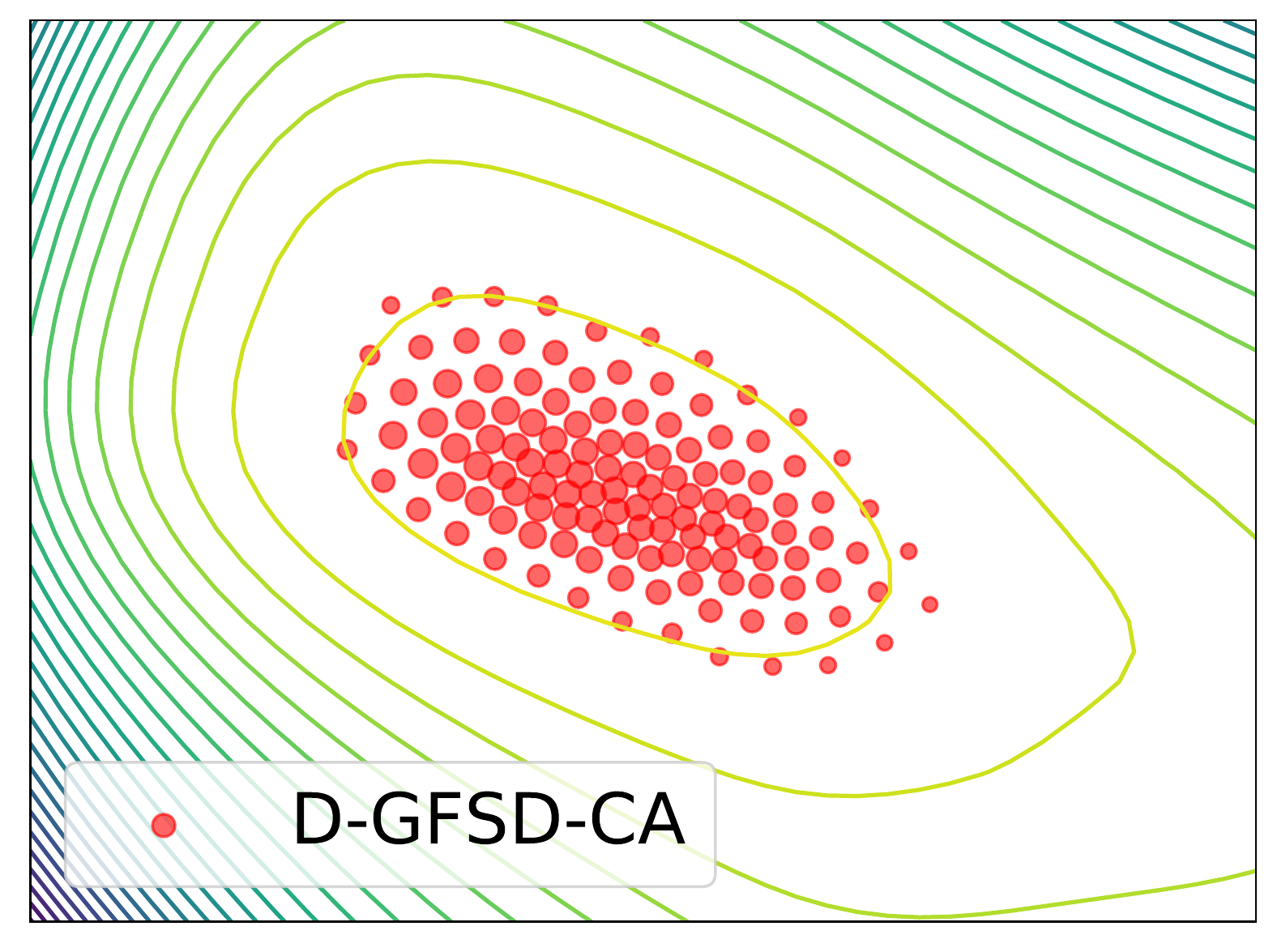}}
	\quad 
	
	\vspace{-4mm}
	\subfigure{\includegraphics[width=.33\columnwidth]{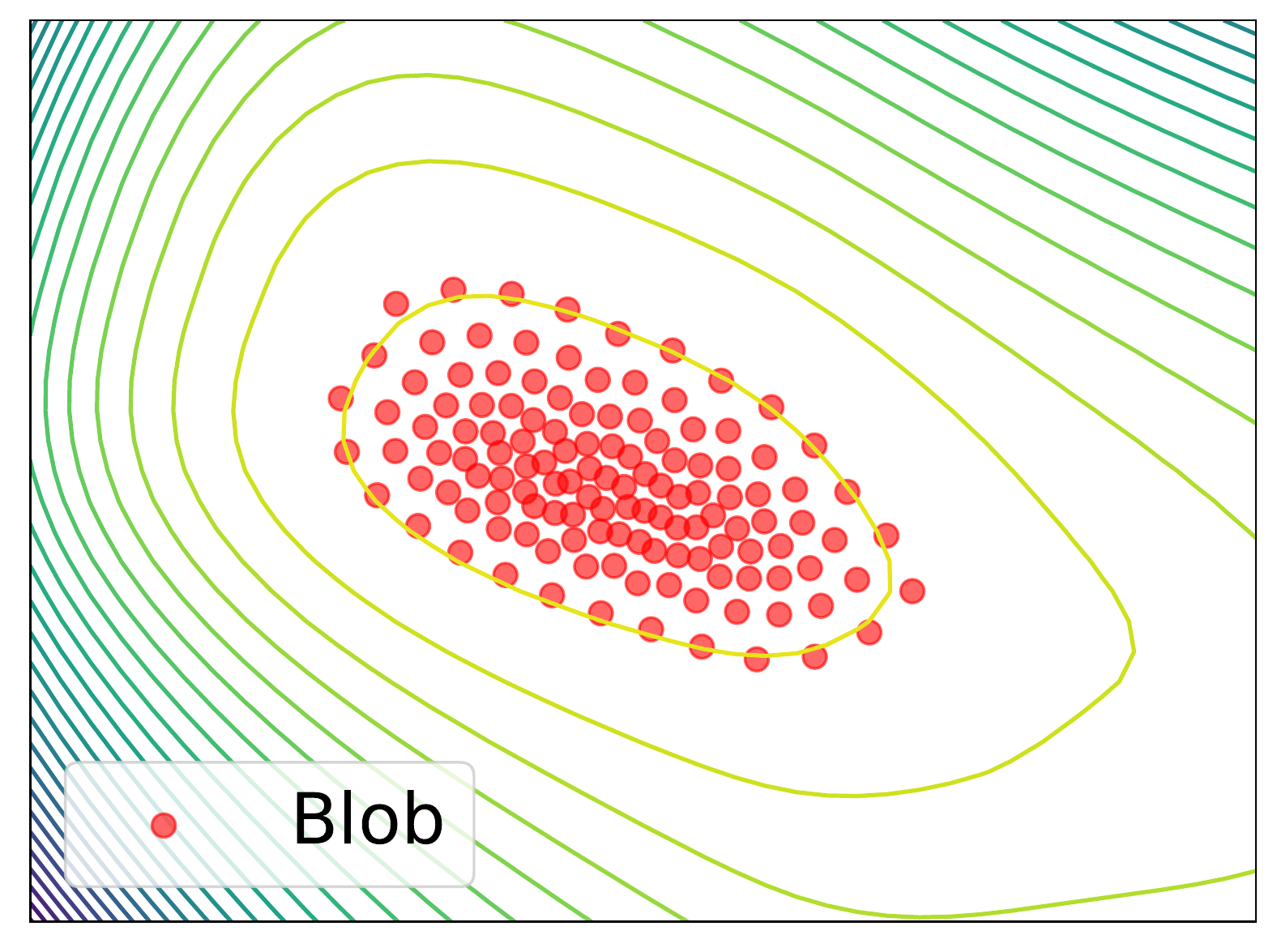}}\hfill
	\subfigure{\includegraphics[width=.33\columnwidth]{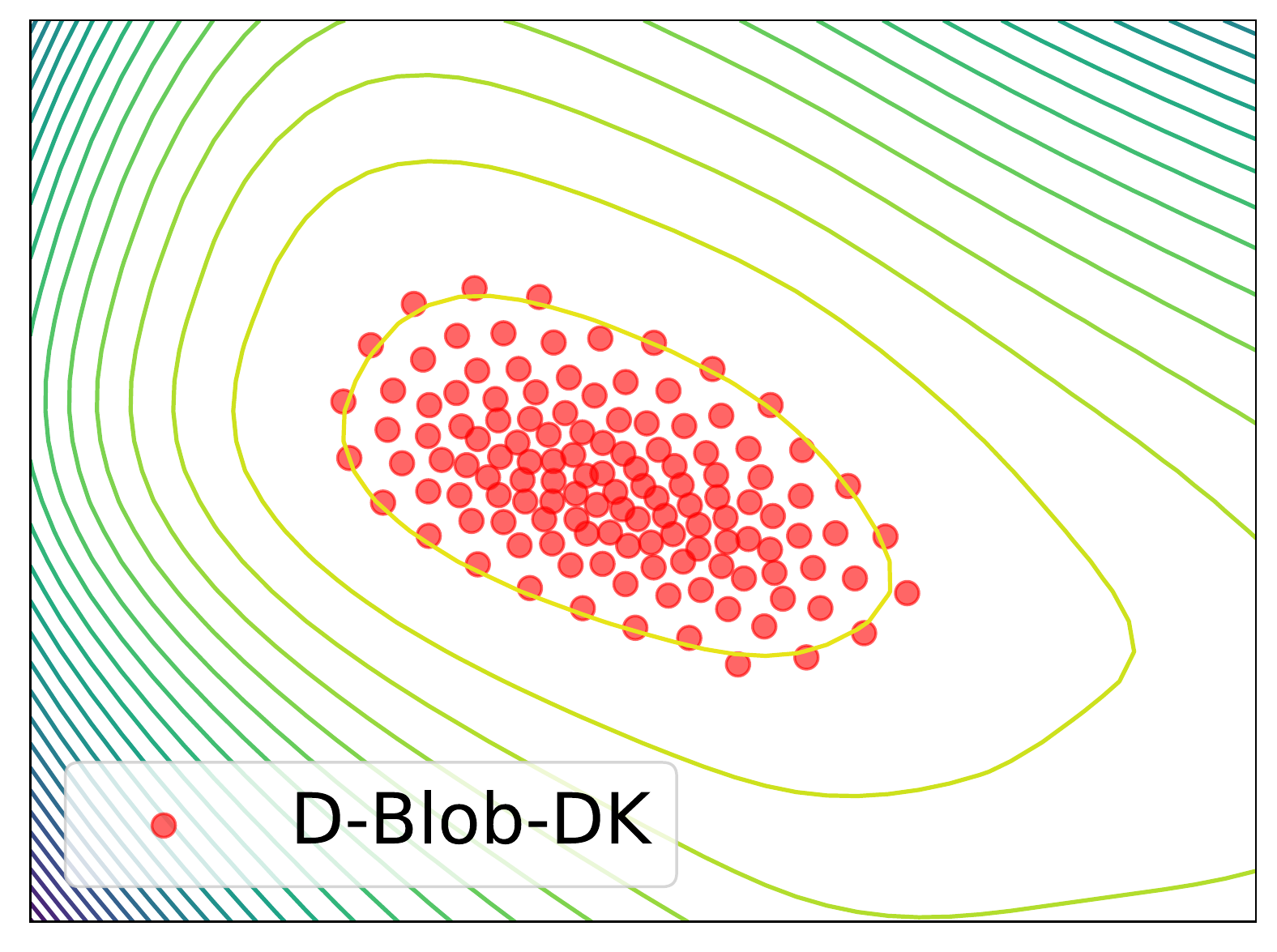}}\hfill
	\subfigure{\includegraphics[width=.33\columnwidth]{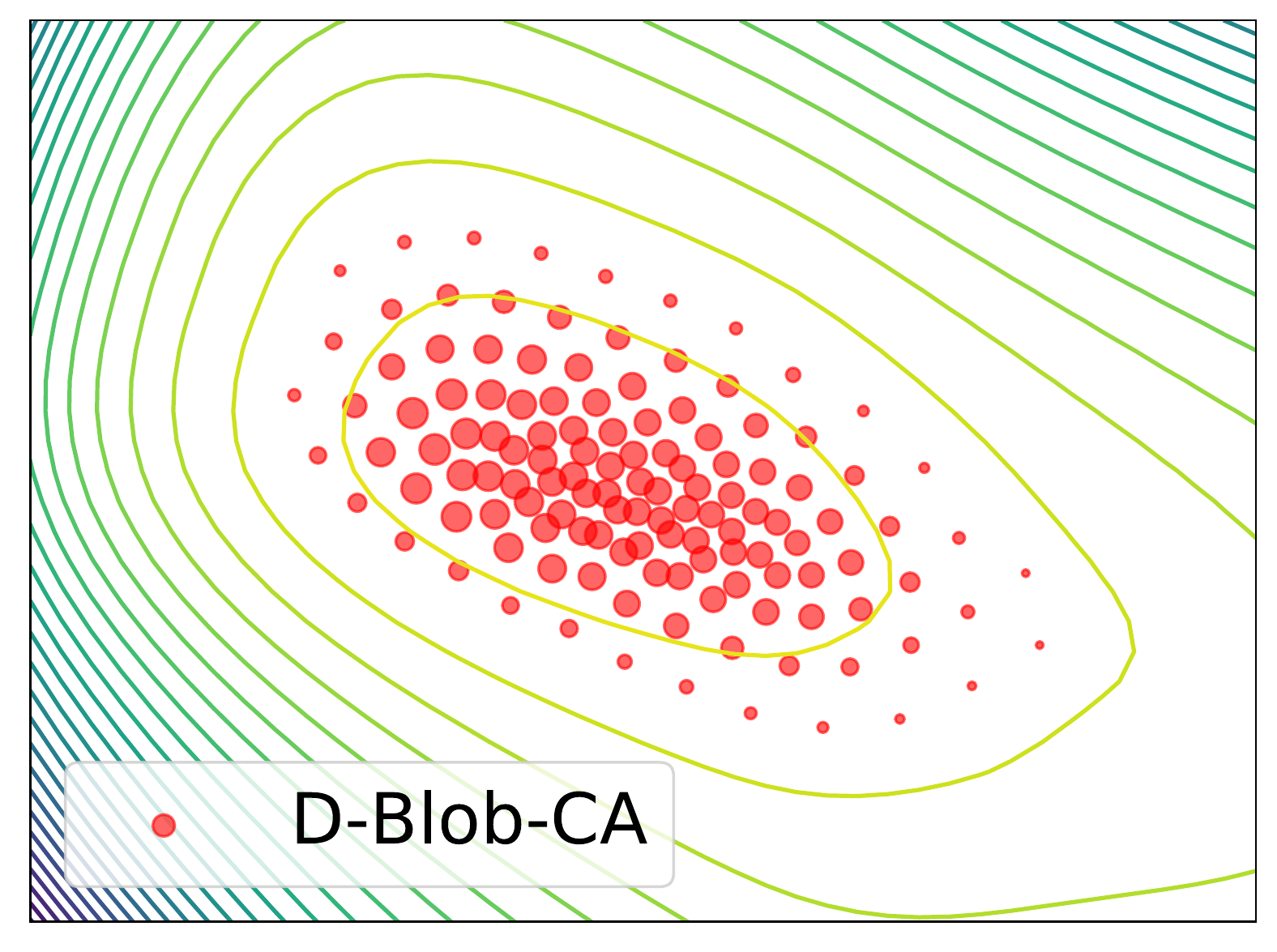}}
	\quad 
	
	\vspace{-4mm}
	\subfigure{\includegraphics[width=.33\columnwidth]{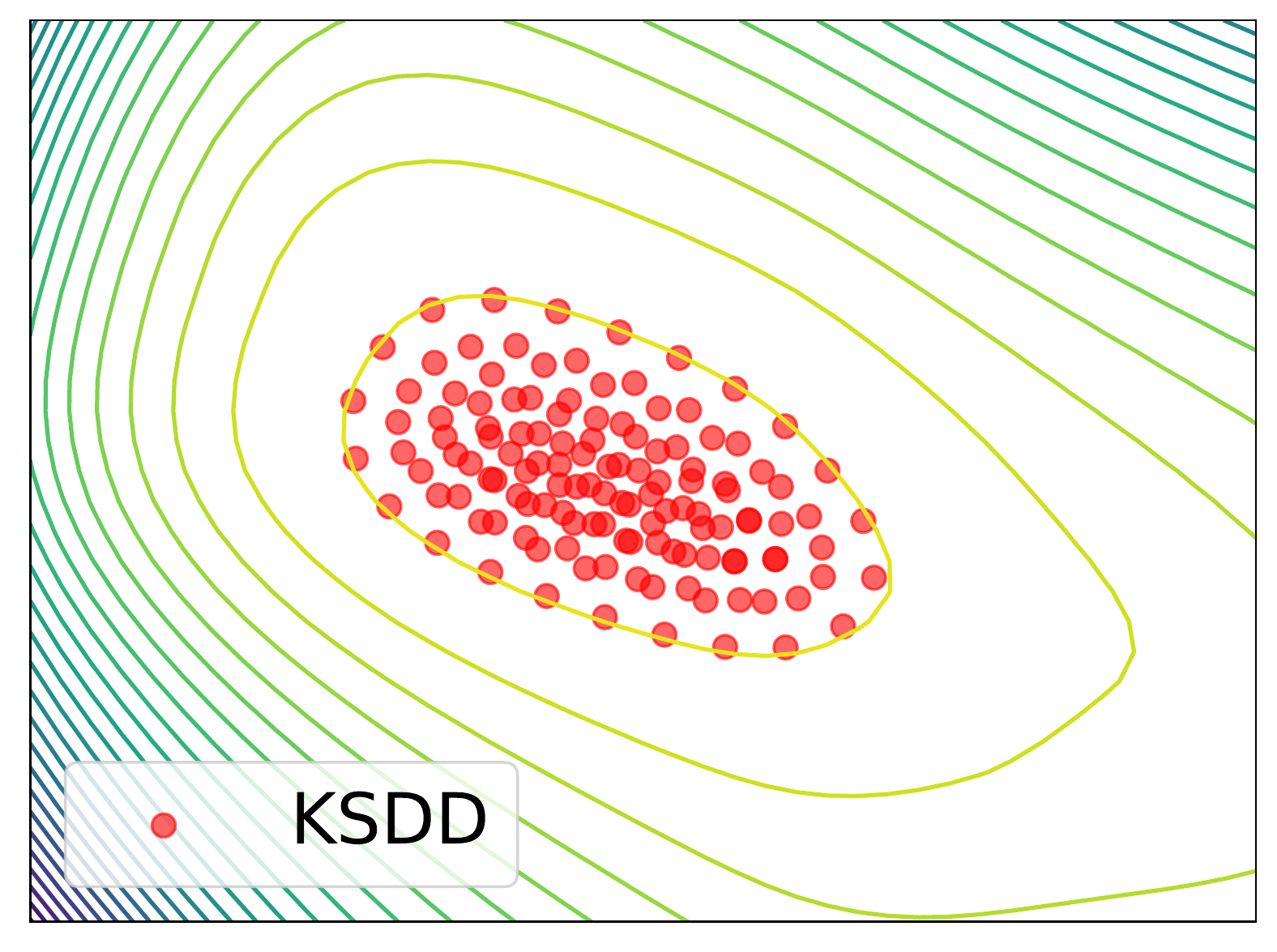}}\hfill
	\subfigure{\includegraphics[width=.33\columnwidth]{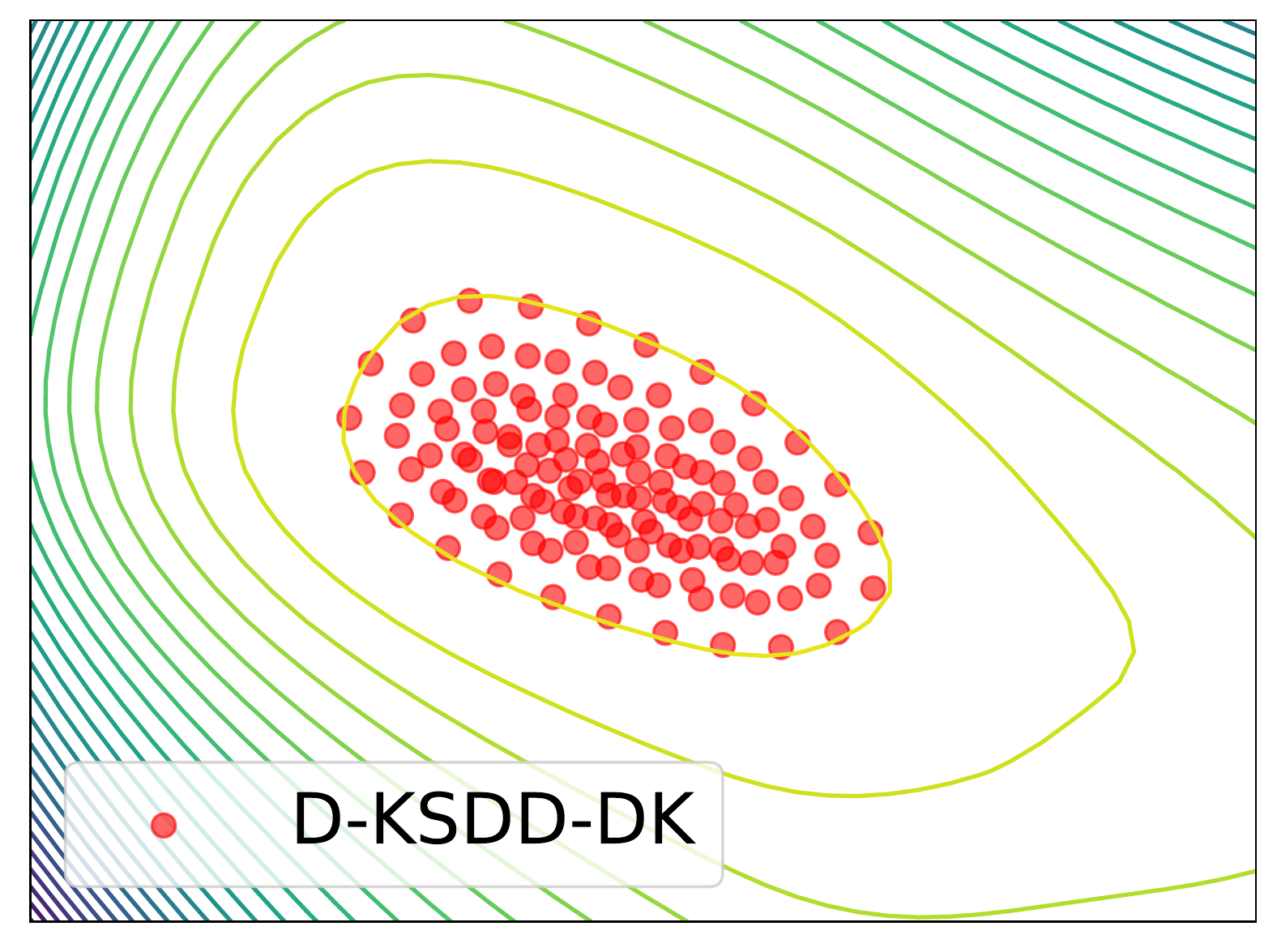}}\hfill
	\subfigure{\includegraphics[width=.33\columnwidth]{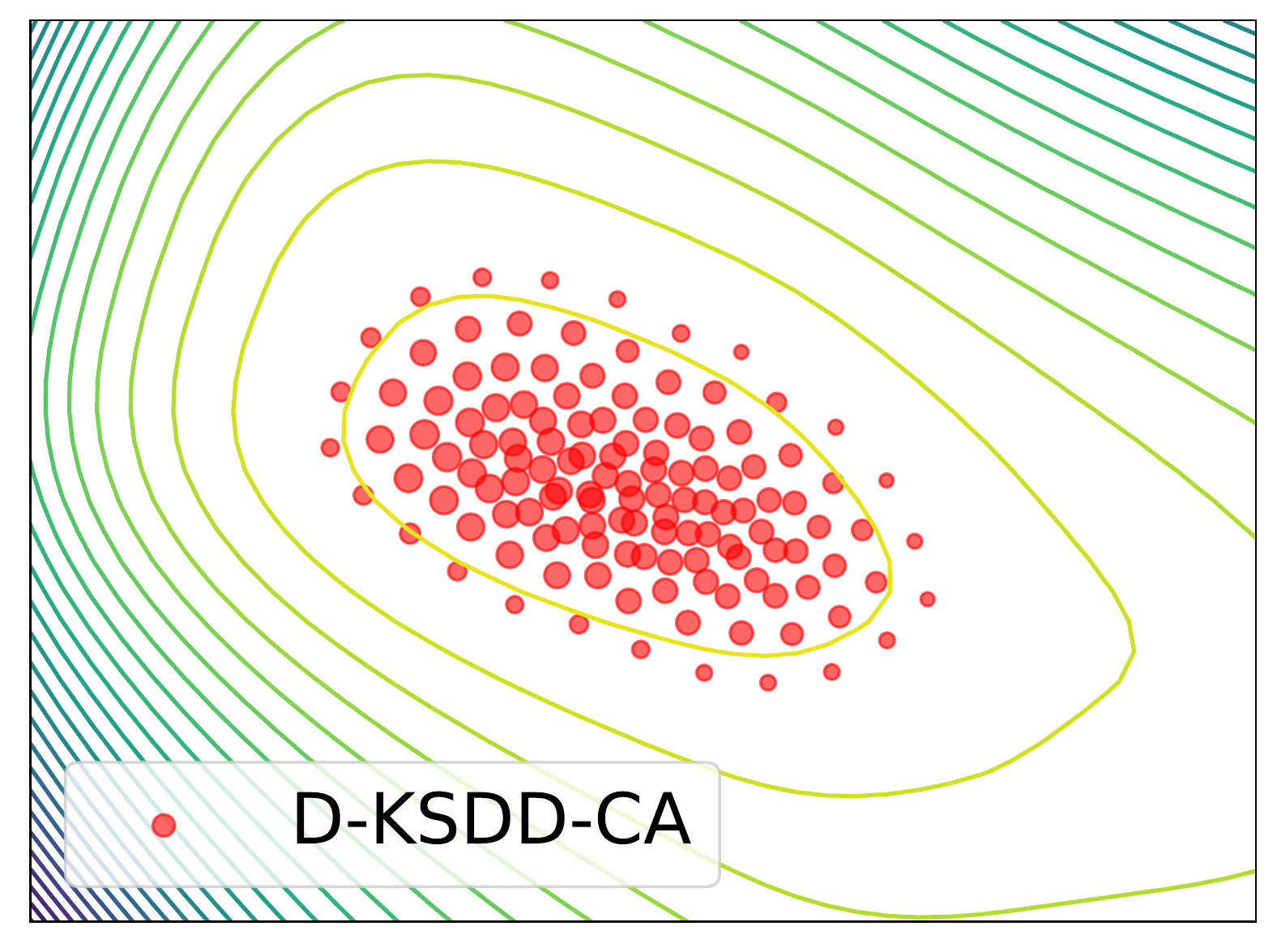}}
	\vspace{-2mm}
	\caption{\label{samples_gp} Approximation results in GP with 128 particles.}
	\vspace{-4mm}
\end{figure}

Figure \ref{samples_gp} gives the contour line of $\log{p(\phi|\mathcal{D})}$, and particles generated by each algorithm.
It is shown that DPVI algorithms with CA achieve better approximation results compared to other algorithms. 
We can also observe that D-Blob-CA has the best performance and covers a wider range of area due to both the dynamic weight adjustment strategy 
	and extra repulsive term.
We note that the DK strategy barely brings any benefits as there is only one mode in this task 
	and both ParVI with fixed-weight and the DK variants of DPVI spread the equally-weighted particles out over this mode.
Actually, the DK variants perform even worse than their fixed-weight ParVI counterparts,
since the duplicate/kill operation induces extra fluctuations in this single-mode model.

To evaluate how well the particles approximate the posterior $p(\phi|\mathcal{D})$, we consider two metrics related with the approximation quality.
Specifically, we report the $W_2$ distance and KSD between the empirical distribution and the target distribution in Table \ref{gp_w2_ksd}, given 
	10000 reference particles generated by the ``gold standard'' HMC method \cite{brooks2011handbook}.
The result demonstrates the effectiveness of our proposed CA weight strategy: DPVI algorithms with CA constantly outperform their fixed-weight counterparts, 
	and the D-Blob-CA algorithm achieves the highest approximation accuracy among all the algorithms.

\subsection{Bayesian Neural Network}
In this experiment, we study a Bayesian regression task with Bayesian neural network on 4 datasets from 
UCI\footnote{http://archive.ics.uci.edu/ml/datasets.php} and LIBSVM\footnote{https://www.csie.ntu.edu.tw/$\sim$cjlin/libsvmtools/datasets/}.
We follow the experiment setting from \cite{liu2016stein}, 
	which models the output as a Gaussian distribution and uses a $\text{Gamma}(1, 0.1)$ prior for the inverse covariance.
We use a one-hidden-layer neural network with 50 hidden units and maintain 128 particles. 
For all the datasets, we set the batchsize as 128.
Since KSDD-type algorithms require evaluating the Hessian matrix of the objective function, which will induce enormous computational burden in 
neural network based tasks, we exclude them in this task.

We report the Root Mean Squared Error (RMSE) of each algorithm in Table \ref{rmse_nn}.
The results show that both the CA and DK weight strategies contribute to a lower RMSE, and DPVI algorithms with CA
	achieve the best performance.
It can be observed that GFSD-type algorithms obtain similar results as Blob-type algorithms, which may be ascribed to the negative correlation between the magnitude of
	the repulsive force and the dimensionality in ParVIs \cite{zhuo2018message}.

\begin{table}[t]
	\centering
	\setlength{\tabcolsep}{5.0mm}{
	\begin{tabular}{c|cc}
		\toprule
		\multirow{2}*{Algorithm} & \multicolumn{2}{c}{Metrics}  \\
		~    & $W_2$ distance & KSD \\
		\hline
		ULD     & 2.365E-1 & 6.838E-2 \\
		BDLS & 2.300E-1 & 6.177E-2 \\
		SVGD    & 1.471E-1 & 6.200E-4 \\
		\hline
		GFSD    & 2.096E-1 & 5.358E-2 \\
		D-GFSD-DK& 2.124E-1 & 5.498E-2 \\
		D-GFSD-CA  & 1.569E-1 & 2.510E-2 \\
		\hline
		Blob    & 1.494E-1 & 3.311E-3 \\
		D-Blob-DK& 1.528E-1 & 3.926E-3 \\
		D-Blob-CA  & \textbf{1.195E-1} & \textbf{5.095E-4} \\
		\hline
		KSDD    & 1.837E-1 & 1.279E-2 \\
		D-KSDD-DK& 1.859E-1 & 1.367E-2 \\
		D-KSDD-CA  & 1.496E-1 & 4.659E-3 \\
		\bottomrule
	\end{tabular}}
	\vspace{-2mm}
	\caption{$W_2$ distance and KSD in GP task.}
	\label{gp_w2_ksd}
	\vspace{-3mm}
\end{table}

\begin{table}[tb]
	\centering
	\resizebox{0.45\textwidth}{!}{
		\begin{tabular}{c|cccc}
			\toprule
			\multirow{2}*{Algorithm} & \multicolumn{4}{c}{Datasets}  \\
			~   & Electrical & Concrete & Kin8nm  & WineRed \\
			\hline
			ULD     & 8.392E+0 & 6.310E+0 & 7.857E-2 & 6.455E-1 \\
			BDLS & 8.378E+0 & 6.310E+0 & 7.843E-2 & 6.452E-1 \\
			SVGD    & 8.717E+0 & 6.189E+0 & 8.044E-2 & 6.403E-1 \\
			\hline
			GFSD    & 8.285E+0 & 6.182E+0 & 7.880E-2 & 6.384E-1 \\
			D-GFSD-DK& 8.226E+0 & 6.179E+0 & 7.866E-2 & 6.384E-1 \\
			D-GFSD-CA  & \textbf{8.058E+0} & 6.173E+0 & \textbf{7.832E-2} & \textbf{6.374E-1}\\
			\hline
			Blob    & 8.286E+0 & 6.181E+0 & 7.880E-2 & 6.384E-1 \\
			D-Blob-DK& 8.226E+0 & 6.178E+0 & 7.866E-2 & 6.384E-1 \\
			D-Blob-CA  & \textbf{8.058E+0} & \textbf{6.171E+0} & 7.834E-2 & \textbf{6.374E-1}\\
			\bottomrule
	\end{tabular}}
	\vspace{-1mm}
	\caption{Averaged Test RMSE.}
	\label{rmse_nn}
	\vspace{-5mm}
\end{table}

\section{Conclusion}
In this paper, we propose a general Dynamic-weight Particle-based Variational Inference (DPVI) framework, which maintains a set of weighted particles 
    to approximate a given target distribution $\pi$ and updates both the particles' positions and weights iteratively. 
Our DPVI framework is developed by discretizing a novel composite flow, which is a combination of a finite-particle approximation of a reaction flow and
    the finite-particle position transport approximation adopted in existing fixed-weight ParVIs.  
We show that the mean-field limit of the proposed composite flow is actually the gradient flow of certain dissimilarity functional $\FM$ in the 
    Wasserstein-Fisher-Rao space, which leads to a faster decrease of $\FM$ than the Wasserstein gradient flow 
    underlying existing fixed-weight ParVIs.
We provide three effective DPVI algorithms with different finite-particle approximations (D-GFSD-CA, D-Blob-CA, and D-KSDD-CA) 
    and derive three variants of them by using the duplicate/kill strategy.
The empirical results show that the proposed DPVI algorithms constantly outperform their fixed-weight counterparts, 
    and D-Blob-CA usually obtains the best performance.

\bibliography{ref}



\end{document}